\documentclass[journal]{IEEEtran}
\usepackage{soul}
\usepackage{gensymb}
\usepackage{balance}
\usepackage{orcidlink}
\usepackage{hyperref}
\usepackage{comment}
\usepackage{booktabs}
\usepackage{caption}
\usepackage{multirow}
\usepackage{graphicx}
\usepackage[export]{adjustbox}
\usepackage[caption=false]{subfig}
\ifCLASSINFOpdf
\else
\fi 
\usepackage{amsmath}
\usepackage{blindtext}
\begin{document}

\title{
Detection of Christmas tree plantations from high-resolution aerial imagery.\\
A case study in the French Morvan
}

\author{Francesca Razzano \orcidlink{0009-0006-0755-0066},~\IEEEmembership{Student Member,~IEEE},
        Emanuele Dalsasso \orcidlink{0000-0001-7170-9015},~\IEEEmembership{Member,~IEEE},
        Adrien Baysse-Lainé \orcidlink{0000-0002-8148-3483},
        Silvia Liberata Ullo \orcidlink{0000-0001-6294-0581},~\IEEEmembership{Senior Member,~IEEE,}
        Gilda Schirinzi \orcidlink{0000-0002-9656-2969},~\IEEEmembership{Senior Member,~IEEE}
        and~\\Jocelyn Chanussot \orcidlink{0000-0003-4817-2875},~\IEEEmembership{Fellow Member,~IEEE}
\thanks{F. Razzano and G. Schirinzi are with Engineering Department, University of Naples Parthenope, Naples, Italy, email: francesca.razzano002@studenti.uniparthenope.it and \nobreak gilda.schirinzi@uniparthenope.it}\nobreak 
\thanks{S. L. Ullo is with the Engineering Department, University of Sannio, Benevento, Italy, email: ullo@unisannio.it}
\thanks{A. Baysse-Lainé is with CNRS Délégation Alpes: Grenoble, Auvergne-Rhône-Alpes, France, email: adrien.baysse-laine@univ-grenoble-alpes.fr}
\thanks{E. Dalsasso and J. Chanussot are with INRIA, Grenoble, France, email: emanuele.dalsasso@inria.fr, jocelyn.chanussot@inria.fr}} \nobreak 

\maketitle

\begin{abstract}

Christmas tree plantations are economically relevant, yet  
a largely unexplored application domain in Remote Sensing (RS).
Their delineation is challenging because of high planting density, short rotation cycles, visual confusion with surrounding vegetation, the availability of dense labels for one reference year only, and severe class imbalance at the landscape scale. Although Deep Learning (DL) methods have shown strong potential for vegetation mapping, existing approaches are typically designed for forests, generic plantation systems, or orchards, and do not explicitly address the structural specificity and hard-negative confusion that characterize Christmas tree plantations. In response to these challenges, this work makes three main contributions: (i) it frames Christmas tree plantation mapping as a distinct rare-target semantic segmentation problem; (ii) it introduces a Hard Negative Mining (HNM) strategy to improve discrimination against confusing background patterns; and (iii) it evaluates the proposed framework across complementary levels, including supervised testing, temporal transfer, and large-scale validation. On the 2020 test set held out, the best model, DeepLabV3 with a ResNet-34 encoder, achieves an IoU of 0.733 and an F1-score of 0.846. HNM substantially improves precision-recall behavior, increasing the area under the precision-recall curve from 0.204 to 0.913. Temporal inference further shows meaningful transferability, reaching IoU/F1 values of 0.751/0.858 on 2017/2018 and 0.691/0.817 on 2023. Large-scale validation further highlights the intrinsic difficulty of the task, as Christmas tree plantations occupied only a very small fraction of the extent of the common evaluation, corresponding to 1,498.4 ha (1.72\%) in 2017/2018 and 1,782.2 ha (2.04\%) in 2023 out of 87,309.4 ha in total. Overall, the proposed framework provides a first step toward the systematic RS of specialized, small-scale vegetation systems that remain poorly represented in current Earth Observation (EO) studies.
\end{abstract}


\IEEEpeerreviewmaketitle

\section{Introduction}

\IEEEPARstart{F}{orests} and tree-based production systems play a central role in global environmental and socio economic dynamics, providing carbon storage, biodiversity habitats, soil protection, and a wide range of ecosystem services that underpin rural economies \cite{10750500, 11348094}. As they concentrate environmental pressures related to land-use change, chemical inputs, and water use, the monitoring of managed tree systems has therefore become a key requirement for climate policy, biodiversity conservation, and sustainable rural development \cite{11154978, XU2026100231}. Within this context, the variety of existing Remote Sensing (RS) systems coupled with automatic image analysis tools can enable to survey such areas and follow their evolution thanks to repeatable observations at appropriate spatial and temporal scale \cite{11173266}. In this study, we focus on the specific case of Christmas tree farming and propose a Deep Learning (DL) pipeline to detect Christmas tree plots from high-resolution aerial imagery in the Morvan Regional Nature Park, France. Christmas tree farming represents a specific form of perennial crop production that lies at the intersection of agriculture (more precisely horticulture) and forestry. Although it occupies a relatively small share of total agricultural land, Christmas tree cultivation has expanded markedly over recent decades and now constitutes a visible and sometimes contested component of rural landscapes in several north-western European regions \cite{baysse2024morvan}. In France, Christmas tree production has been at the center of debates over its ecological footprint, with the Morvan Regional Nature Park frequently cited as one of the main producing regions where dedicated plantations reshuffle local land-use patterns.
This expansion has raised increasing concerns about environmental impacts and land-use competition \cite{isabelle2013, bayasse2024}. In the Morvan, environmental organisations and local collectives have pointed to the effects of monoculture plantations and agrochemical use on soil quality, water resources, and biodiversity, triggering public debate and regulatory scrutiny. At the same time, producers highlight the economic importance of the sector and ongoing efforts to reduce chemical inputs and adopt more sustainable practices, for instance through eco-responsibility labels and protected geographical indication schemes. These tensions underscore the need for reliable, spatially explicit information on where Christmas tree plantations are located, how extensive they are, and how they evolve over time, as a basis for environmental assessment, land-use planning, and socio-political analysis.
While the synergy between Earth observation data (offering wall-to-wall coverage at high spatial resolution) and automatic image analysis tools (with modern AI-based vision systems ensuring scalability and efficiency) has transformed the way land uses are analyzed \cite{zhao2023systematic}, digital tools for operational practices in Christmas tree farms are still missing \cite{streitberger2021phytodiversity}. Indeed, existing methods are typically developed and evaluated on natural or managed forests, generic plantation forestry, or fruit orchards, without considering the specific spatial layout, management cycles, and socio environmental relevance of Christmas tree farming. Given the economic importance, environmental footprint, and public visibility of Christmas tree cultivation, the lack of dedicated studies suggests that there is still a methodological and application gap to be addressed in the RS and DL literature.  

\begin{figure*}[ht]
    \centering
    \includegraphics[width=1\textwidth]{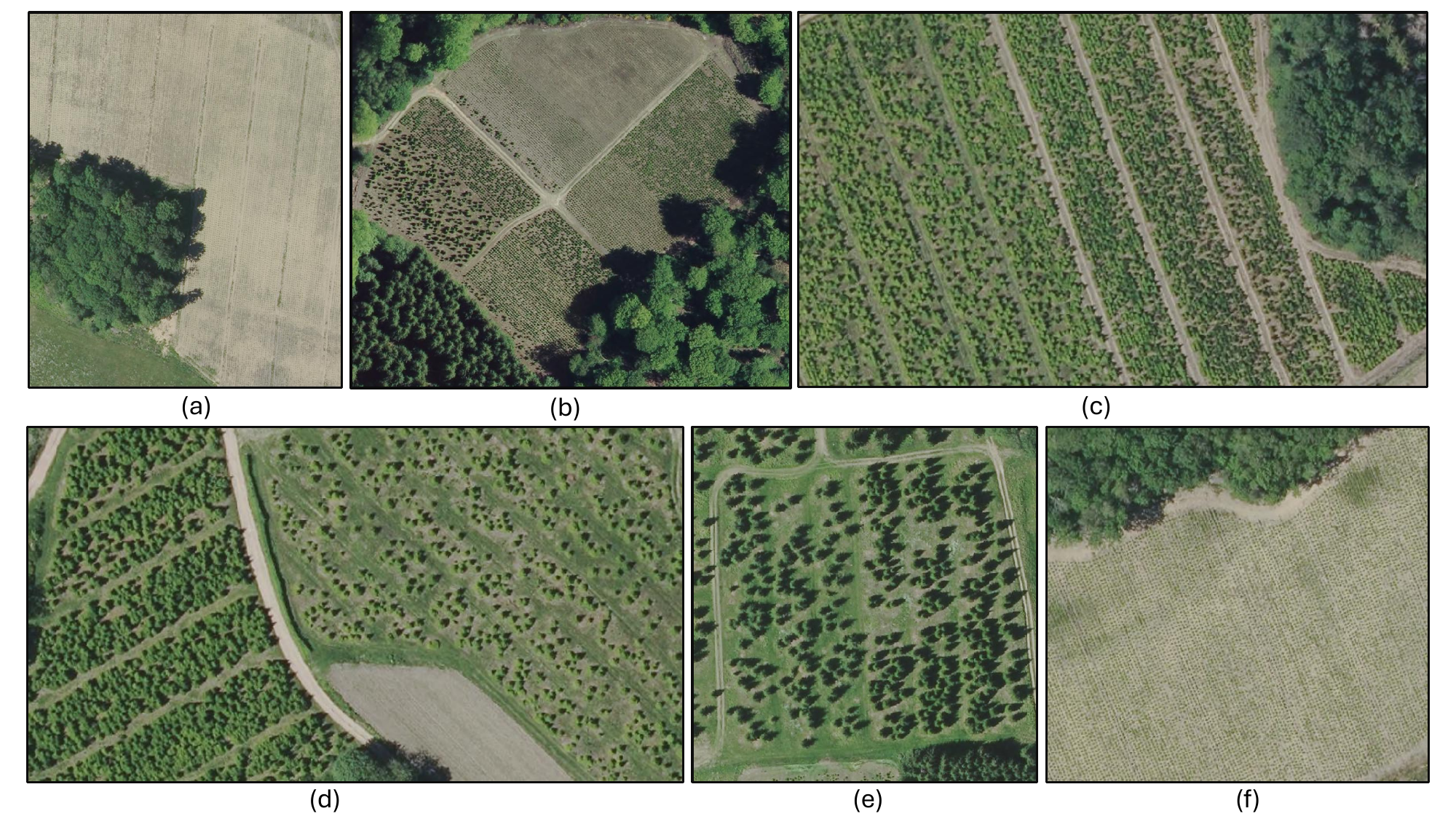} 
    \caption{Examples of the high visual and structural variability of Christmas tree plantations in high-resolution RGB imagery. Despite belonging to the same target class, parcels exhibit drastic differences depending on their management phase and spatial context: (a) newly established plantation on bare soil; (b) variable density and row spacing across different sub-plots; (c) mature and densely planted parcel with strong linear patterns; (d) irregular harvesting and clearing dynamics; (e) sparse canopy cover with grassy background; (f) early-stage planting adjacent to forest borders.}
    \label{fig:plantation_variability}
\end{figure*}

In this context, the present study addresses Christmas tree plantation mapping as a dedicated EO task and investigates how DL-based semantic segmentation can be adapted to this highly specific setting. Indeed, Christmas tree plantations cannot be treated as a simple extension of either orchard mapping or conventional forest monitoring. From a RS and DL perspective, they define a particularly challenging target class for at least three reasons. First, they exhibit a highly specific structural and temporal signature, combining very high planting densities, often reaching 8,000--10,000 trees per hectare, with short harvest rotations of 5--10 years, which makes them fundamentally different from both orchards and forests. 
Consequently, their visual appearance is extremely variable and depends strongly on the specific management stage. As illustrated in Fig. \ref{fig:plantation_variability}, parcels labeled as active plantations can range from bare soil or early-stage planting to fully mature and dense canopies, often presenting irregular clearing patterns, internal grassy strips, or variable spacing. This intra-class heterogeneity makes them fundamentally different from both regular mature orchards and homogeneous forests.
Second, their visual appearance may strongly overlap with that of other land-cover types, including grasslands, shrub encroachment, recent clear-cuts, and young reforestation, especially when only RGB imagery is available. Third, these difficulties are compounded by scarce labeling and severe class imbalance at landscape scale, as plantation areas occupy only a very small fraction of the total area under analysis \cite{baysse2024morvan}. To answer the highlighted challenges, the main contributions of this work are threefold: (i) we formalize Christmas tree plantation delineation as a distinct RS challenge, clearly differentiating it from both forests and orchards in terms of planting density, management dynamics, and visual appearance; (ii) we design a dedicated hard-negative-aware dataset construction, training, and evaluation pipeline, including multi-class mask generation, semantically guided hard negative sampling, and a loss formulation tailored to severe class imbalance and visually confusing backgrounds;  (iii) we deploy our model at the scale of the main production area, covering a large share of the Natural Regional Park -- while groundtruth labels are available for a single year only (2020), we propose an automatic labeling strategy that enables to assess model's accuracy at different years (namely 2017 and 2023).

\section{Related works}
\label{sec:rel_work}

RS has long supported the monitoring of managed perennial crops and tree-based production systems. Early studies in precision viticulture highlighted the value of optical sensing for characterizing canopy vigor, spatial variability, and crop condition, thereby establishing a broader methodological foundation for image-based monitoring of intensive woody crops \cite{https://doi.org/10.1111/j.1755-0238.2002.tb00209.x}. More recent works further confirmed the relevance of multi-platform observations for vineyard analysis and management, combining satellite, Unmanned Aerial Vehicles (UAV), and ground information to capture spatio-temporal variability \cite{pastonchi2020comparison}. In parallel, recent studies on vineyard abandonment and parcel monitoring have shown the growing importance of AI-based image analysis for operational agricultural management, both from aerial imagery and from multi-temporal satellite data \cite{TEIXEIRA20242038,Roussel01072026}. These studies demonstrate that remotely sensed data can support the monitoring of specialized perennial crops, but they mainly focus on viticulture and parcel classification rather than on dense conifer plantations with highly specific visual patterns.

\textbf{A first relevant line of research concerns the semantic segmentation of agricultural plots and cultivated tree canopies from high-resolution imagery}.
In the context of parcel delineation, Qi et al. \cite{rs16020346} compared several deep architectures, including U-Net, SegNet, DeepLabV3+, and TransUNet, and showed that transformer-enhanced encoder--decoder models can improve agricultural plot segmentation performance. At finer spatial scales, Li et al. \cite{LI2024108538} proposed a lightweight improved U-Net for fruit tree canopy segmentation from UAV orthophotos, combining attention mechanisms and focal loss to improve efficiency and accuracy in densely planted orchards. Similarly, Zhang et al. \cite{10.3389/fpls.2022.1041791} addressed instance segmentation of orchard canopies through an enhanced Mask R-CNN framework, with particular emphasis on precise boundary delineation under branch overlap and shadow effects. More recently, Sun et al. \cite{11024197} proposed a feature fusion mechanism based on YOLOv8 to improve the instance segmentation of tiny and densely distributed tree crowns in rural aerial imagery, demonstrating the necessity of integrating multi-scale features for complex canopy layouts. Taken together, these studies confirm the effectiveness of DL for agricultural and orchard segmentation, but they primarily target regular canopy objects or well-structured cultivated layouts, which differ substantially from the characteristics of Christmas tree plantations.

\textbf{A second stream of literature focuses on individual tree detection and crown delineation in forest and heterogeneous tree-covered environments.} Weinstein et al. \cite{rs11111309} demonstrated that semi-supervised DL can detect individual tree crowns from RGB imagery even when manually annotated data are limited, an important result for RS tasks with scarce labels. This direction was extended to heterogeneous forests by Beloiu et al. \cite{beloiu2023individual}, to urban forestry applications by Zhang et al. \cite{9765982}, and to very high resolution satellite imagery by Korznikov et al. \cite{f12010066}, who showed that U-Net-like architectures can outperform more conventional pixel-based classifiers in precise tree recognition. Building on this foundation, recent studies have further optimized encoder-decoder networks for canopy extraction; for instance, Duan et al. \cite{11087559} proposed an attention-enhanced multi-scale U-Net (CIA-UNet) to improve single tree crown segmentation by dynamically weighting spatial features and resolving boundary adhesion issues. 
Beyond individual studies, the review by Diez et al. \cite{rs13142837} highlighted the growing maturity of DL methods based on UAV-acquired RGB data for practical forestry applications, including tree detection, species classification, and anomaly analysis. More recently, Jarahizadeh and Salehi \cite{JARAHIZADEH2026115088} proposed a dedicated DL architecture for individual tree detection from UAV LiDAR-derived rasters, further illustrating the rapid methodological progress in tree-level monitoring. Similarly addressing complex canopy features, recent advancements have focused on robust boundary delineation and domain shifts; for instance, Wu et al. \cite{11424282} introduced a dual-level domain alignment and boundary-aware optimization framework for semi-supervised tree crown instance segmentation in UAV imagery. While these works mostly focus on individual crowns or forest stands, in this study we target the delineation of managed plantation parcels whose internal texture is dense, repetitive, and often visually confusable with surrounding vegetation.

\textbf{A third line of work addresses large-scale land-cover and vegetation monitoring using multi-temporal satellite observations.} In this context, the combination of Sentinel-1 and Sentinel-2 time series has proven to be particularly effective for land-cover mapping, thanks to the complementarity between radar and optical information \cite{IENCO201911}. Parcel-level agricultural monitoring has also been explored through targeted workflows that combine satellite markers, field observations, and complementary in situ image acquisition, as shown by d'Andrimont et al. \cite{rs10081300} for grassland monitoring. More broadly, grasslands have become an important application domain in RS, both as a target class and as a source of confusion in agricultural and ecological mapping problems \cite{rs14122903, He31122026}. Related datasets and operational products have also become increasingly important for supporting vegetation analysis in France, including reference data from natural grassland \cite{PANHELLEUX2023109348} and submonthly clear-cut monitoring systems derived from Sentinel-1 time series \cite{10604724}. 
\begin{figure*}[ht]
    \centering
    \includegraphics[width=0.78\textwidth]{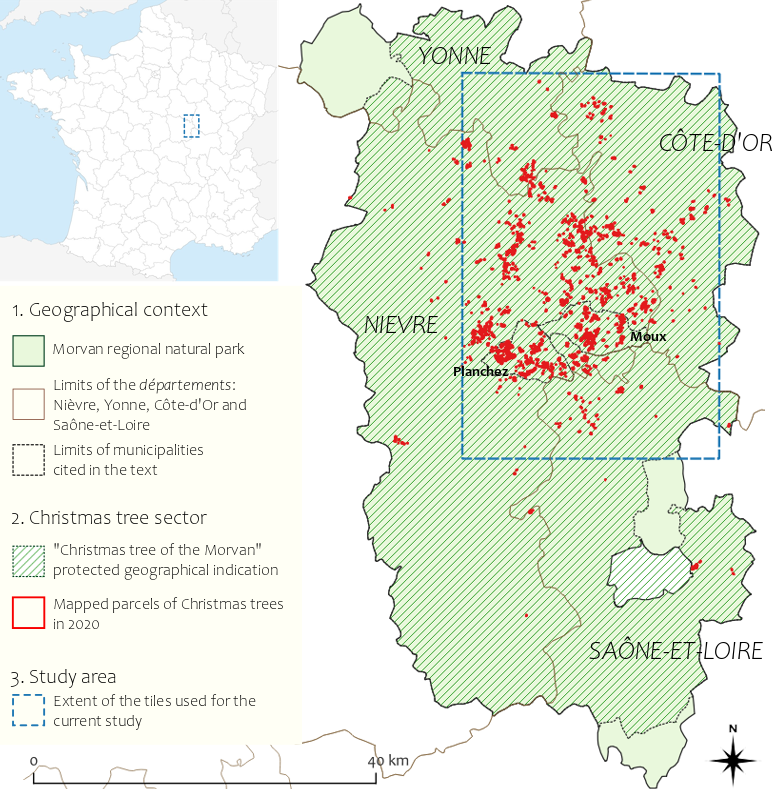} 
    \caption{Study area in the Morvan Regional Natural Park, France. The figure shows the geographical context of the Morvan area within France, the extent of the protected geographical indication ``Christmas tree of the Morvan'', the parcel-level inventory of Christmas tree plantations mapped in 2020 (red), and the spatial extent of the orthophoto tiles used in this study (blue dashed rectangle).}
    \label{fig:morvan_coverage}
\end{figure*}

\textbf{A complementary line of research addresses semantic segmentation and object detection under severe class imbalance}.
In this latter case, rare targets are embedded in large, visually heterogeneous backgrounds. In computer vision, hard example mining strategies have been introduced to increase the contribution of ambiguous or misclassified samples during optimization, thereby improving discrimination in challenging settings \cite{shrivastava2016ohem}. In parallel, focal loss was proposed as an effective alternative to explicit sample selection by down-weighting easy examples and concentrating training on hard negatives, especially in highly imbalanced dense prediction problems \cite{lin2017focal}. These ideas are particularly relevant in RS, where target classes may occupy only a small fraction of the image and where visually structured backgrounds can generate a large number of false positives. In this context, Zhang et al. \cite{zhang2020hardexample} proposed a hard-example-mining network for object detection in high-resolution RS imagery, while Jemaa et al. \cite{jemaa2023tree} used a hard negative mining strategy to improve UAV-based tree detection under complex background conditions. Their results support the idea that explicitly modeling difficult non-target examples can substantially improve robustness in vegetation-rich scenes. In our case, we follow the same general rationale, but adapt it to the specific problem of Christmas tree plantation delineation by explicitly structuring the training set around semantically meaningful hard negatives, including grasslands, clear-cuts, and other visually confusable non-target areas.
In this work, we draw inspiration from these methodological contributions to address the specific problem of monitoring Christmas tree plantations.

\section{MATERIAL FOR STUDY AREA}
\label{sec:dinsar}

\subsection{Study area: The Morvan region, France}
The study is conducted in the Morvan Regional Natural Park, a low-mountain massif located in central France at the northern edge of the Massif Central. This area provides a particularly relevant test bed for Christmas tree plantation monitoring, as it represents the main production hub in France and has been the subject of a detailed spatial inventory of plantations in 2020 by Baysse-Lainé and Bernard \cite{baysse2024morvan}, which we use here as the reference Ground Truth (GT) for our analysis. 
More specifically, the 2020 inventory constitutes the only dense parcel-level GT reference available for Christmas trees and therefore serves as the main supervised basis for model training and quantitative evaluation.

Morvan region constitutes an interesting study area as it presents the following features: 
\begin{itemize}
    \item according to recent geographical studies, the Morvan supplies approximately one-third of the national Christmas tree demand, making it not only an important agricultural area but also a territory where plantation expansion has visible landscape effects \cite{baysse2024morvan}. 
    \item The location of plantation parcels reflects a compromise between topographic suitability (a substantial share of Christmas tree cultivation is concentrated between 560 and 680~m \cite{baysse2024morvan}), production constraints and mechanization requirements (producers favor moderate slopes), and climatic suitability: although cooler northern exposures might appear theoretically preferable, the observed distribution of parcels shows a predominance of south-west, south, and west-oriented slopes.
    \item The production system is highly concentrated spatially, resulting in both dense hotspots of plantations and more isolated peripheral parcels, which are challenging to model from a RS perspective. 
    \item The seasonal rhythm of field operations, with harvesting concentrated in late autumn, contributes to the temporal variability of parcel appearance. Moreover, active cultivation areas shift over time; in parallel, abandoned parcels may gradually evolve toward dense monospecific wooded formations. This makes the distinction between active plantations, recent abandonment, young reforestation, and surrounding vegetation increasingly difficult. 
\end{itemize}

\begin{table*}[t]
\centering
\caption{Summary of the main datasets used in this study, their temporal coverage, and their role in the workflow.}
\label{tab:data_summary}
\renewcommand{\arraystretch}{1.75}
\begin{tabular}{p{2.9cm}p{1.9cm}p{1.4cm}p{2.7cm}p{1.8cm}p{5.2cm}}
\hline
\textbf{Data} & \textbf{Year(s)} & \textbf{Resolution} & \textbf{Source} & \textbf{Reference} & \textbf{Role in workflow} \\
\hline
BD ORTHO orthophotos 
& 2017/2018, 2020, 2023 
& 0.2 m 
& IGN 
& \cite{IGN_Descriptif_2022} 
& Main RGB imagery for supervised training (2020), temporal inference (2017/2018 and 2023), and visual interpretation. \\

Manual parcel annotations 
& 2020 
& Parcel level 
& Based on expert photo-interpretation and field verification 
& \cite{bayasse2024}
& Positive reference layer for Christmas tree plantations; used for training, validation, and test construction. \\

OSO land-cover map 
& 2018, 2020 
& 10 m 
& THEIA 
& \cite{Inglada2017,theia_landcover} 
& Water masking for 2017/2018 and 2020; exclusion of obvious non-target areas. \\

CLC+ Backbone 
& 2023 
& 10 m 
& Copernicus Land Monitoring Service 
& \cite{Copernicus_CLCplus} 
& Water masking for the 2023 temporal inference dataset. \\

Natural grassland dataset 
& 2020 
& 10 m 
& Panhelleux et al. 
& \cite{PANHELLEUX2023109348} 
& Grassland-related hard negative support for 2020; identification of herbaceous areas likely to be confused with Christmas tree plantations. \\

HRL Grassland 
& 2017, 2023 
& 10 m 
& Copernicus Land Monitoring Service 
& \cite{copernicus_grassland_2017,copernicus_grassland_2023} 
& Grassland support layers for temporal inference in 2017/2018 and 2023. \\

Clear-cut layers 
& 2017--2023 
& 10 m 
& Mermoz et al. 
& \cite{10604724} 
& Hard negative mining, temporal filtering, and exclusion of recently harvested forest parcels potentially confused with plantation areas. \\
\hline
\end{tabular}
\end{table*}
This temporal turnover is one of the main reasons why a static inventory, although extremely valuable as a reference baseline, is insufficient for long-term monitoring. A continuous RS approach is therefore needed to capture both the persistence and the transformation of cultivated areas in this evolving landscape.

\subsection{Data and layers for GT}

To train and validate the proposed monitoring system, we relied on a high-quality reference dataset centered on the Morvan region. The GT dataset focuses on the core production area within the Morvan Regional Natural Park, particularly on the municipalities where Christmas tree cultivation is most concentrated, such as Moux-en-Morvan and Planchez. 
As shown in Fig. \ref{fig:morvan_coverage}, the study area extends across four administrative units equivalent to provinces or counties (NUTS 3 level in the EU classification), called départements in France, with most of the data located in the Ni\`evre, which corresponds to the main production nucleus.
To guarantee complete coverage of the annotated parcels, we selected a total of 54 high-resolution orthophoto tiles from the Institut national de l’information géographique et forestière (IGN) database. Their distribution mirrors the spatial clustering of Christmas tree production across the four departments: 33 tiles are located in the Ni\`evre, which hosts the densest western and central production hubs, 11 in C\^ote-d'Or, 7 in Sa\^one-et-Loire, and 3 in Yonne. This sampling strategy was designed to capture the full diversity of the landscape, ranging from the dense plantation clusters of the production core to more isolated parcels situated in peripheral areas. Overall, the dataset provides the reference basis for developing and evaluating our RS framework, while its components play different roles across the workflow. High-resolution orthophoto imagery is available for three temporal snapshots, namely 2017--2018, 2020, and 2023. Among these, only the 2020 acquisition is initially paired with a rigorous vector GT, which serves as the main supervised reference for model training, validation, and first-stage evaluation. 
In contrast, the 2017–2018 and 2023 orthophotos were first used as inference targets to investigate the temporal transferability of the model beyond the reference year. For quantitative and qualitative evaluation, however, a subset of these data was also manually annotated, enabling performance assessment on the available reference areas.
Finally, additional GT information is established for 2017--2018 and 2023 to support a large-scale temporal evaluation of the detected Christmas tree cultivation areas, which remain a rare target class compared with the surrounding landscape. 
Ancillary layers, including grassland, clear-cut, and water masks, are integrated throughout the workflow with two complementary purposes: in the 2020 setting, they are used to identify hard negative samples and improve class discrimination during training; in the 2017--2018 and 2023 settings, they are further used to exclude implausible non-target areas, constrain the search space, and support the construction of temporally consistent reference data for large-scale evaluation. All the main datasets used in this study are reported in Table \ref{tab:data_summary}:

\subsubsection{Orthophoto Imagery and Temporal Series}
The primary visual input of this study is provided by the BD ORTHO\textsuperscript{\textregistered} database, i.e., the official high-resolution orthophotographic product distributed by the French National IGN \cite{IGN_Descriptif_2022}. To support both supervised learning and temporal generalization analysis, we assembled three orthophoto mosaics representative of the study area at different time steps: 2017/2018, 2020, and 2023. All orthophotos share a common ground sampling distance of 20 cm, which ensures geometric consistency and visual comparability across years. However, because IGN acquisition campaigns are organized at the departmental level, each yearly mosaic is not strictly synchronous over the entire Morvan study area. In particular, the earliest snapshot corresponds to a composite built from acquisitions performed in 2017 and 2018 across the four departments intersecting the Morvan Regional Natural Park (Ni\`evre, C\^ote-d'Or, Yonne, and Sa\^one-et-Loire), while the 2020 and 2023 mosaics were similarly assembled from the corresponding departmental campaigns \cite{IGN_Descriptif_2022}.
Within this temporal framework, the 2020 orthophotos constitute the main reference imagery used for supervised model development, as they are paired with manually verified parcel annotations. By contrast, the 2017/2018 and 2023 orthophotos are primarily used as target years for temporal inference and change analysis, allowing us to investigate the spatial evolution of Christmas tree plantations before and after the reference year.

\begin{figure*}[ht]
    \centering
    \includegraphics[width=18cm]{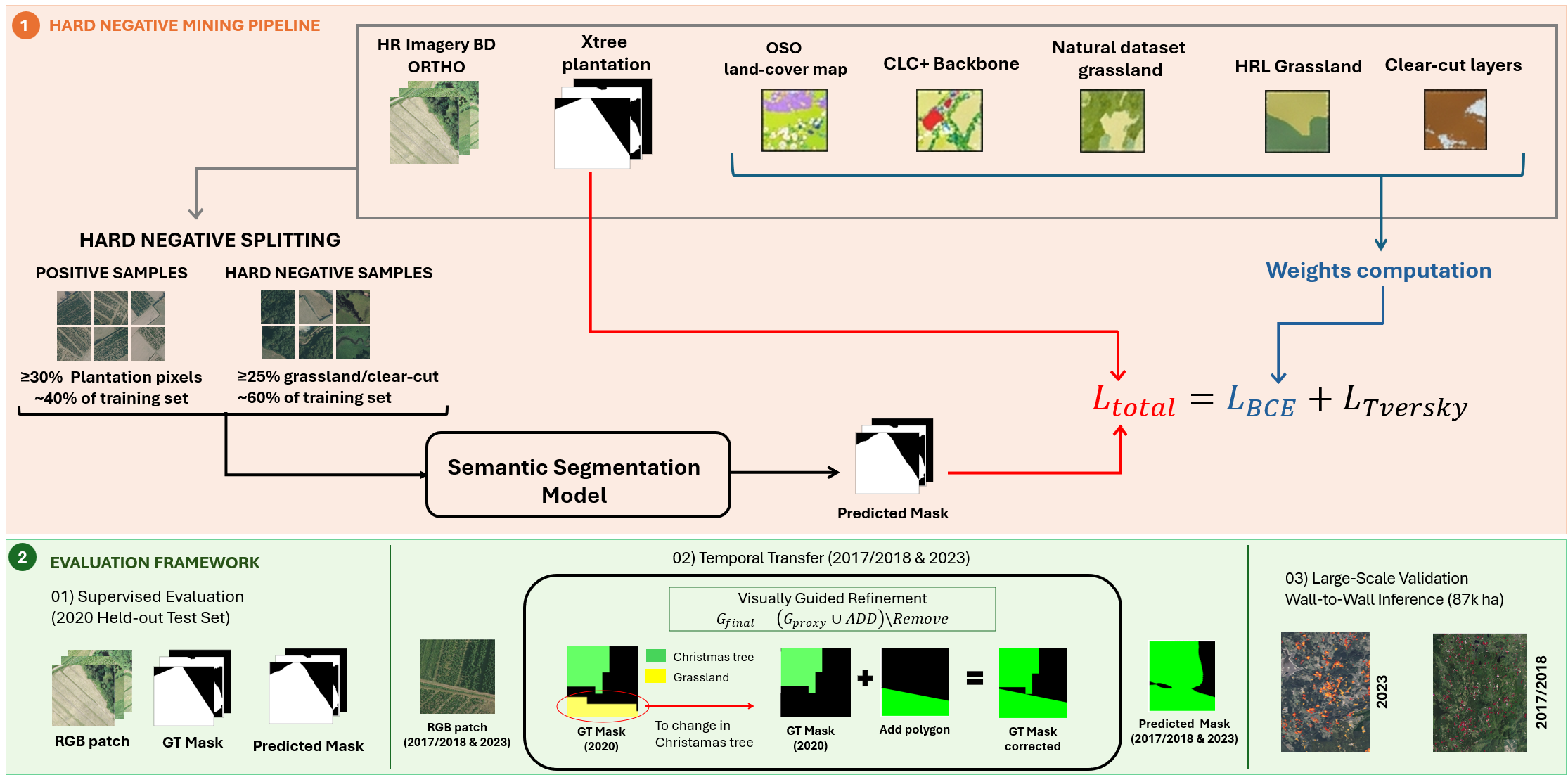}
    \caption{Overview of the proposed workflow for Christmas tree plantation delineation from high-resolution RGB imagery. The pipeline combines BD ORTHO orthophotos, parcel-level reference data, and ancillary land-cover layers to construct a hard-negative-aware training dataset. Positive and hard negative samples are then used to train the semantic segmentation model, which is evaluated both on the 2020 held-out test set and in temporal transfer settings over 2017/2018 and 2023 through visually guided refinement and large-scale analysis.}
    \label{fig:workflow}
\end{figure*}

\subsubsection{Ground Truth Vector Data}
The supervised reference layer consists of 1,253 manually annotated parcels describing active Christmas tree plantations in 2020. This vector dataset was originally constructed through a rigorous multi-stage photo-interpretation and field validation protocol designed to discriminate Christmas tree plantations from visually similar land covers, including young forest plantations, fallow parcels, and recently disturbed areas \cite{baysse2024morvan}.
More specifically, the annotation workflow combined: i) an initial parcel screening based on LPIS information and visual inspection of orthophotos; ii) a geometric refinement of parcel boundaries directly on the 2020 imagery; iii) temporal cross-checking against historical imagery to resolve ambiguous cases, particularly in areas potentially confused with clear-cuts or early-stage reforestation; iv) a criteria-based visual filtering phase; and v) field verification and discussions with local stakeholders and producers.
A parcel was retained as a valid Christmas tree plantation only when it exhibited a consistent set of morphological indicators visible in the orthophotos. These indicators included regular planting geometry, high within-parcel tree density, and characteristic management features such as internal grass strips (\textit{tourni\`eres}) or sharp parcel boundaries marked by fences or maintenance contrasts. This filtering strategy was essential to minimize label noise and to ensure that the positive class represented active and managed production sites rather than generic conifer cover.

\subsubsection{Ancillary Data and Temporal Support Layers}
\label{subsub:ancillary}
Because Christmas tree plantations may be confused with other low- to medium-height vegetated surfaces, we complemented the orthophotos and parcel annotations with ancillary raster layers describing water bodies, grassland-related areas, and clear-cuts. 
These ancillary layers were primarily used to identify difficult negative contexts for training and temporal validation. In particular, they helped characterize background surfaces such as grassland-related areas, clear-cuts, and water-adjacent contexts that may exhibit visual patterns potentially confusable with Christmas tree plantations, thereby supporting a more challenging and representative definition of negative samples across time.
Grassland-related information was included because herbaceous parcels and managed grasslands frequently generate hard negatives, particularly in young or sparse plantation phases. For 2020, we used the natural grassland dataset proposed by Panhelleux et al. \cite{PANHELLEUX2023109348}, which provides a 10~m raster map over mainland France derived from multi-year land-cover consistency and accompanied by ground reference information. For the temporal inference years, grassland support layers were obtained from the Copernicus High Resolution Layer Grassland products, using the 2023 release for the 2023 orthophotos and the 2017 release as the closest available reference for the 2017/2018 imagery period \cite{copernicus_grassland_2023,copernicus_grassland_2017}.
Finally, clear-cut layers were incorporated to reduce confusion with recently harvested forest parcels, which may locally resemble newly established or recently cleared Christmas tree fields. For this purpose, we used the clear-cut products released by Mermoz et al. \cite{10604724}, which provide temporally explicit disturbance information over mainland France and explicitly use ancillary layers in their own workflow to support masking and analysis. In our case, these layers were used as temporal support masks and as a source of hard negative candidates, both for the 2020 supervised setting and for the 2017/2018 and 2023 inference analyses.

\subsection{Dataset Construction}
Fig.~\ref{fig:workflow} summarizes the overall workflow adopted in this study for the dataset construction. Starting from RGB orthophotos, parcel annotations, and ancillary layers, the pipeline first constructs a hard-negative-aware dataset, then trains the segmentation model, and finally evaluates it in both supervised and temporal transfer settings.
Building a robust training dataset from very high-resolution orthophotos is challenging because Christmas tree plantations occupy only a very small fraction of the total Morvan landscape. As a result, a naive patch extraction strategy would produce a dataset largely dominated by non-target background areas, with only limited exposure to the target class. To address this issue, we designed a dedicated dataset construction pipeline aimed at i) increasing the representation of verified Christmas tree parcels, ii) explicitly preserving difficult non-target contexts, and iii) avoiding spatial leakage during model evaluation.

\subsubsection{Patch Extraction and Multi-Class Mask Construction}

The original orthophoto tiles, each covering approximately $5 \times 5$ km ($25{,}000 \times 25{,}000$ pixels at 0.2 m ground sampling distance), were too large to be directly processed by the segmentation models. Each tile was therefore subdivided on a regular non-overlapping grid, producing image patches of $512 \times 512$ pixels, corresponding to about $102.4 \times 102.4$ m on the ground. Across the full set of selected tiles, this patch extraction stage produced approximately 122,000 candidate patches before the subsequent sampling step. For each image patch, a corresponding multi-class semantic mask was generated by combining the manual parcel annotations with the ancillary raster layers introduced in Section \ref{subsub:ancillary}. The resulting masks distinguish five categories: background, Christmas tree plantations, grassland, clear-cuts, and water. In this formulation, the Christmas tree class is derived from the manually verified parcel layer, whereas the remaining classes are used to retain structured information on non-target land-cover types that may otherwise be merged into a generic background category. This design was motivated by the fact that several non-target surfaces may still resemble Christmas tree plantations in RGB orthophotos, especially in terms of texture, spatial organization, or overall visual appearance. In particular, grassland areas and recently disturbed forest parcels may generate false positives if treated as undifferentiated background. Preserving these categories explicitly in the mask therefore allowed us to encode semantically meaningful hard negatives already at the dataset construction stage.

\subsubsection{Mask Integration and Label Priority}

Because the ancillary layers originated from multiple external products with coarser spatial resolution than the orthophotos, their integration with parcel-level annotations required particular care. The final patch masks were generated by combining the available layers according to a fixed label priority. Christmas tree parcels were assigned first, followed by grassland, clear-cut, and water classes only where no higher-priority label had already been assigned. In practice, this means that pixels belonging to verified Christmas tree parcels retained priority over ancillary classes, while the remaining labels were added sequentially to structure the non-target space.

\begin{figure}[htbp]
    \centering
    \includegraphics[width=0.5\textwidth]{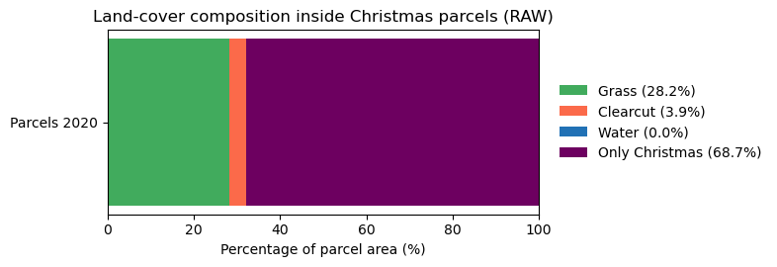} 
    \caption{Land-cover composition inside the verified Christmas tree parcels (2020 dataset). The bar chart illustrates the raw spatial overlap between the manually annotated parcels and the ancillary layers (grassland, clear-cut, and water), motivating the use of explicit priority rules during multi-class mask generation.}
    \label{fig:landcover_composition}
\end{figure}

To assess the potential impact of class overlap, we also quantified the raw spatial intersections between the verified Christmas tree parcels and the ancillary layers. As shown in Fig.~\ref{fig:landcover_composition}, most of the parcel area was correctly associated with the plantation class (68.7\%), but substantial overlaps with grassland (28.2\%) and smaller overlaps with clear-cuts (3.9\%) were also observed before filtering. These results confirm that direct fusion of all layers would have introduced non-negligible label ambiguity, justifying the use of explicit label-priority rules during mask generation.

\subsubsection{Hard Negative Sampling and Dataset Split}

Once the candidate patches and their associated masks had been generated, we applied a selective sampling strategy to avoid constructing a dataset dominated by easy background examples. Rather than sampling patches uniformly, we retained representative positive samples and deliberately enriched the dataset with difficult negative contexts.

A patch was considered a positive sample when Christmas tree plantations covered at least 30\% of its pixels. This threshold ensured that the model was trained on sufficiently representative plantation patterns rather than on sparse or highly marginal occurrences. To avoid over-representing a small number of highly productive areas, the number of positive samples was capped at $P_{cap}=150$ patches per tile.

Negative samples were mined more selectively. Only patches containing no Christmas tree pixels were considered eligible as negatives, and they were retained only when they contained at least 25\% of either grassland or clear-cut area. This strategy focused the negative set on visually and texturally confusing contexts rather than on arbitrary background. In addition, water was preserved as a separate class in the multi-class masks, and a limited number of water-containing patches could also be retained among the negative candidates when available, in order to preserve non-target diversity.
The final number of negative patches selected per tile was dynamically controlled relative to the number of positives, with a target negative-to-positive ratio of $r_{target}=2.0$ and an upper cap of $N_{cap}=150$ negative patches per tile. This sampling strategy allowed the dataset to remain balanced enough for training while still emphasizing difficult negative examples.
Finally, to avoid spatial autocorrelation between training and evaluation data, the dataset was partitioned at the tile level rather than at the patch level. This ensured that no patches extracted from the same orthophoto tile appeared in more than one subset. The final partition followed an approximate 70\%/15\%/15\% split for training, validation, and test, respectively, computed over the full set of tiles. This tile-level partition provided a more rigorous basis for evaluating generalization over unseen geographic areas.

\section{METHODOLOGY}
\subsection{Model Training Setup}
\label{loss_sec}
To maximize the robustness of the model given the limited and imbalanced nature of the dataset, we implemented specific strategies for data loading, augmentation, and loss calculation.\\

\subsubsection{Data Augmentation Strategy}
Since the dataset consists of a limited number of annotated patches, data augmentation was adopted to reduce overfitting and improve invariance to geometric and radiometric variability. We implemented a custom on-the-fly augmentation pipeline (\textit{RandomAugment}) during training, combining random horizontal and vertical flips ($p=0.5$ each), random rotations by multiples of $90^\circ$ ($p=0.75$), brightness jitter within $\pm 20\%$ ($p=0.5$), contrast jitter within $\pm 30\%$ relative to the image mean ($p=0.5$), and additive Gaussian noise with standard deviation $\sigma=0.02$ ($p=0.3$). Such transformations are well suited to aerial imagery, where scene orientation is arbitrary and local radiometric variability may affect appearance. To preserve label consistency, geometric transformations were applied jointly to the image and the segmentation mask, whereas radiometric perturbations were applied only to the image.

\subsubsection{Loss Function}
To address the dual challenge of extreme class imbalance and visual confusion with similar land covers, we designed a hybrid objective function named \textit{HardNegativeLoss}. This objective combines weighted Binary Cross-Entropy (BCE), which emphasizes pixel-wise discrimination, with Tversky loss, which promotes overlap-based consistency and is explicitly configured to prioritize recall of the minority class. The overall loss is defined as
\begin{equation}
    \mathcal{L}_{total} = \lambda_{bce} \mathcal{L}_{BCE} + \lambda_{tversky} \mathcal{L}_{Tversky}
\end{equation}
where the balancing coefficients were empirically set to $\lambda_{bce} = 0.4$ and $\lambda_{tversky} = 0.6$.

The first term, $\mathcal{L}_{BCE}$, incorporates the spatially explicit weight map $W$ directly into the optimization process in order to explicitly guide the model towards learning discriminative features for the positive class against hard negative classes. In particular, by assigning larger penalties to errors occurring in challenging regions such as grasslands, clear-cuts, and water, this component encourages the model to learn fine-grained visual and textural differences between plantations and confusing background classes. Formally, for a batch of $N$ pixels, the weighted BCE term is expressed as
\begin{equation}
    \mathcal{L}_{BCE} = - \frac{1}{N} \sum_{i=1}^{N} w_i \cdot \left[y_i \log(\hat{y}_i) + (1-y_i) \log(1-\hat{y}_i)\right]
\end{equation}
where $y_i \in \{0,1\}$ denotes the binary ground-truth label, $\hat{y}_i \in [0,1]$ the predicted probability, and $w_i$ the class-dependent weight assigned to pixel $i$. The weights were empirically set to penalize errors on critical classes more heavily:
\begin{equation}
    w_i =
    \begin{cases}
    1.0 & \text{if pixel } i \in \text{Background (0)} \\
    4.5 & \text{if pixel } i \in \text{Christmas Tree (1)} \\
    2.0 & \text{if pixel } i \in \text{Grassland (2)} \\
    2.5 & \text{if pixel } i \in \text{Clear-cut (3)} \\
    6.0 & \text{if pixel } i \in \text{Water (4)}
    \end{cases}
\end{equation}
The highest weight assigned to the target class counteracts its scarcity, while elevated weights for grassland and clear-cut areas penalize false positives in difficult hard-negative regions. Water pixels were assigned an even larger penalty to strongly discourage spurious detections over obvious non-target areas. As a result, misclassifications in hard-negative regions contribute more strongly to the optimization than errors over generic background pixels.

The second term, $\mathcal{L}_{Tversky}$, is derived from the Tversky index, a generalization of the Dice coefficient that allows asymmetric control over false positives and false negatives:
\begin{equation}
    TI = \frac{TP}{TP + \alpha FP + \beta FN + \epsilon}
\end{equation}
where $TP = \sum \hat{y}_i y_i$, $FP = \sum \hat{y}_i (1-y_i)$, and $FN = \sum (1-\hat{y}_i) y_i$ denote true positives, false positives, and false negatives, respectively, and $\epsilon$ is a small constant introduced for numerical stability. In our setting, the hyperparameters were fixed to $\alpha=0.2$ and $\beta=0.8$, chosen after preliminary experiments, so that false negatives were penalized more strongly than false positives. This choice reflects the operational objective of the study, namely favoring high recall and limiting missed detections of small or isolated plantation parcels. The corresponding loss term is finally defined as
\begin{equation}
    \mathcal{L}_{Tversky} = 1 - TI .
\end{equation}

By combining weighted BCE and Tversky loss, the proposed objective jointly improves local discrimination over hard negatives and the global structural coherence of the predicted segmentation masks.

\subsubsection{Training Protocol}

Training was conducted for a maximum of 20 epochs using mixed-precision computation (\textit{torch.amp}) to reduce memory usage and accelerate processing. At each iteration, input patches were passed through the network to obtain output logits, the \textit{HardNegativeLoss} was evaluated using the predicted logits, ground-truth masks, and associated pixel-wise weight maps, and gradients were backpropagated to update the model parameters. Model selection was performed on the validation set by monitoring Intersection-over-Union (IoU) after each epoch, and the weights achieving the highest validation IoU were retained as the final \textit{best model} for testing.

\subsection{Model Architecture and Hyperparameters}\label{subsection:model_architecture}

As semantic segmentation model for the detection of Christmas tree plantation, we use a DeepLabV3 with ResNet34 encoder (DeepLabV3-R34). To justify our choice, we compared it within a unified experimental setting against the following baseline architectures: a custom U-Net baseline with base width 16, DeepLabV3 with ResNet-18 and ResNet-50 encoders, Feature Pyramid Network (FPN) with a ResNet-34 encoder, and U-Net++ with a ResNet-34 encoder. All encoder-based architectures were implemented using the \texttt{segmentation\_models\_pytorch} library and initialized with ImageNet-pretrained weights to leverage transfer learning and improve convergence stability.

All models were configured to process three-channel RGB aerial imagery and to produce a single-channel binary output mask representing the presence or absence of Christmas tree plantations. The custom U-Net architecture followed the classical encoder--decoder design with skip connections, using an initial number of 16 base filters that doubled at each downsampling stage. Dropout with a probability of $p=0.3$ was applied within convolutional blocks to reduce overfitting while preserving spatial detail.

\begin{figure*}[t]
\centering
\includegraphics[width=\textwidth]{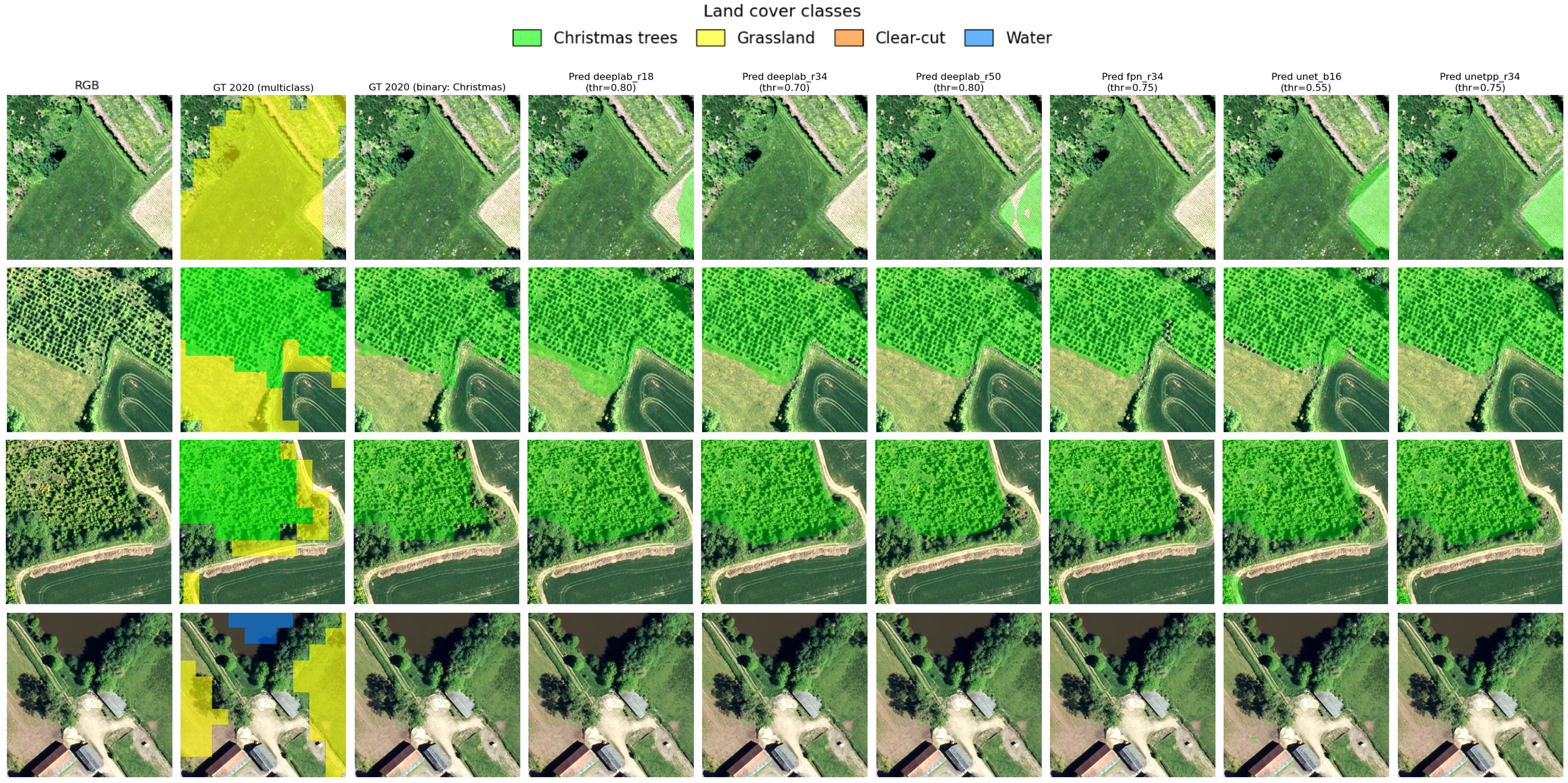}
\caption{Qualitative comparison on the 2020 test set. Columns show (from left to right) RGB patch, GT 2020 multiclass mask, GT binary mask (Christmas tree class), and predictions from DeepLabV3-R18, DeepLabV3-R34, DeepLabV3-R50, FPN-R34, U-Net (b16), and U-Net++-R34 (thresholds as in Table~\ref{tab:metrics_2020}).}
\label{fig:qual_2020_grid}
\end{figure*}

Training was conducted on a CUDA-enabled GPU using mixed-precision (FP16) computation to reduce memory consumption and accelerate training. Optimization was performed using the Adam optimizer with an initial learning rate of $1 \times 10^{-4}$. To ensure stable convergence, a \textit{ReduceLROnPlateau} scheduler was employed, halving the learning rate whenever validation IoU failed to improve for three consecutive epochs.

Because all architectures produced probabilistic output maps, a post-training threshold optimization was performed on the validation set. 
Threshold values in the range [0.05, 0.95] were systematically evaluated, and a single operating threshold was selected for each architecture as the one maximizing the validation F1-score. 
This procedure ensured that binary predictions were not derived from an arbitrary cutoff and allowed performance to be tuned to the specific segmentation objective. Model comparison was conducted on the 2020 test set, and the architecture achieving the best balance between Intersection over Union and F1-score was selected for temporal inference experiments and large-scale deployment.

\subsection{Evaluation Metrics}
Model performance was assessed using global pixel-level metrics computed over the entire test area. Given the strong class imbalance between plantation and non-plantation pixels, we relied primarily on Intersection over Union (IoU) and F1-score, as these metrics are robust to skewed class distributions and directly reflect segmentation quality.

Precision and recall were also reported to characterize the trade-off between false positives and false negatives. While precision quantifies the proportion of predicted plantation pixels that are correct, recall measures the proportion of ground-truth plantation pixels successfully detected. The F1-score was computed as the harmonic mean of precision and recall, providing a balanced summary of detection performance.

In large-scale inference experiments, confusion matrices were accumulated across all patches to compute global metrics over millions of pixels. In addition to IoU and F1-score, specificity and balanced accuracy were reported to better interpret model behavior under extreme class imbalance conditions. Balanced accuracy, defined as the average of recall and specificity, ensures that performance is not artificially inflated by the predominance of background pixels. All reported metrics correspond to binary segmentation results obtained using the validation-optimized threshold.

\section{Results}
\label{results}

We first establish a robust baseline on the 2020 test set (258 samples), where the availability of dense and reliable GT masks enables both quantitative and qualitative analysis. This provides a test bed for the comparison of the selected semantic segmentation models, as described in Section \ref{subsection:model_architecture}. The analysis on the 2020 test set supports the choice of DeepLabV3-R34 as the reference architecture for this study. On the retained DeepLabV3-34, we first ablate on the loss function to underscore the importance of combining a weighted BCE with the Tversky loss, and then study the impact of the proposed HNM strategy.

Finally, we deploy the model on different years by extending the analysis on 2017/2018 and 2023. First, we restrain the temporal analysis to a spatial support corresponding to the 2020 test set: because pixel-level GT was not available for those years, we explicitly describe a manual visual annotation procedure used to extend the GT to 2027/18 and 2023, enabling a consistent quantitative and qualitative assessment across time and the pixel level. Then, we evaluate the scalability of the approach through large-scale inference over the full study area, producing wall-to-wall plantation probability and binary maps and validating them against independent expert annotations.

\subsection{Evaluation on 2020 Test Set}
\label{subsec:supervised test}

\subsubsection{Selection of the semantic segmentation model}
In this section, we benchmark different segmentation architectures under the same HNM-aware training setup. All models are trained using the same HardNegativeLoss, i.e. the hybrid objective combining weighted BCE and Tversky loss described in Section \ref{loss_sec}.
We evaluate the candidate segmentation models on the held-out 2020 test set, which is the only acquisition year where dense, parcel-level GT masks are available. This provides a reliable test bed to evaluate models' generalization to unseen areas and assess model's accuracy in Christmas tree plantation delineations. Probabilistic outputs were binarized using the validation-optimized threshold (reported for each model), and metrics were computed by aggregating pixel-wise confusion matrices over the full test set.
Table~\ref{tab:metrics_2020} summarizes the quantitative comparison. 
\begin{table}[ht]
\centering
\caption{Performance on the 2020 test set. Thresholds (thr) are selected on the validation set by maximizing the F1-score. Best values are highlighted in bold. Arrows indicate whether higher ($\uparrow$) values correspond to better performance.}
\label{tab:metrics_2020}
\begin{tabular}{lccccc}
\hline
\textbf{Model} & \textbf{thr} & \textbf{IoU $\uparrow$} & \textbf{F1 $\uparrow$} & \textbf{Prec. $\uparrow$} & \textbf{Rec. $\uparrow$} \\
\hline
DeepLabV3-R18           & 0.80 & 0.703 & 0.826 & 0.830 & 0.821 \\
\textbf{DeepLabV3-R34}  & \textbf{0.70} & \textbf{0.733} & \textbf{0.846} & \textbf{0.866} & 0.827 \\
DeepLabV3-R50           & 0.80 & 0.697 & 0.822 & 0.807 & 0.837 \\
FPN-R34                 & 0.75 & 0.700 & 0.824 & 0.833 & 0.814 \\
U-Net (b16)             & 0.55 & 0.512 & 0.677 & 0.762 & 0.609 \\
U-Net++-R34             & 0.75 & 0.731 & 0.845 & 0.843 & \textbf{0.846} \\
\hline
\end{tabular}
\end{table}
Among all evaluated architectures, DeepLabV3-R34 achieves the best overall performance, reaching the highest IoU ($0.733$) and F1-score ($0.846$), with a strong precision--recall balance (precision $0.866$, recall $0.827$). U-Net++-R34 follows closely (IoU $0.731$, F1 $0.845$), while lighter backbones (e.g., DeepLabV3-R18) remain competitive but slightly below. The ViT-based U-Net baseline exhibits substantially lower overlap and recall in this setting. Based on these results, DeepLabV3-R34 is selected as the reference model for all subsequent experiments.
\begin{table*}[ht]
\centering
\caption{Multi-seed loss ablation on DeepLabV3-R34 over the 2020 test set with fixed threshold 0.70. Values are reported as mean $\pm$ standard deviation across seeds. Best values are highlighted in bold.}
\label{tab:loss_ablation}
\begin{tabular}{lcccc}
\hline
\textbf{Loss} & \textbf{IoU $\uparrow$} & \textbf{F1 $\uparrow$} & \textbf{Prec. $\uparrow$} & \textbf{Rec. $\uparrow$} \\
\hline
Tversky & 0.577 $\pm$ 0.067 & 0.730 $\pm$ 0.054 & 0.655 $\pm$ 0.053 & 0.826 $\pm$ 0.058 \\
Weighted BCE & 0.667 $\pm$ 0.042 & 0.800 $\pm$ 0.031 & \textbf{0.769 $\pm$ 0.102} & 0.849 $\pm$ 0.081 \\
\textbf{Weighted BCE + Tversky} & \textbf{0.695 $\pm$ 0.009} & \textbf{0.820 $\pm$ 0.006} & 0.756 $\pm$ 0.017 & \textbf{0.895 $\pm$ 0.011} \\
\hline
\end{tabular}
\end{table*}
To complement the quantitative evaluation, Fig.~\ref{fig:qual_2020_grid} provides a qualitative comparison on representative 2020 patches. Overall, the models correctly capture the characteristic plantation textures and block geometry, but they differ in terms of boundary coherence and robustness to confusing backgrounds (e.g., crops, grasslands, and heterogeneous forest edges). In agreement with Table~\ref{tab:metrics_2020}, DeepLabV3-R34 yields the most stable delineations with fewer spurious detections. 

\subsubsection{Ablation on the training objective}
We performed an additional ablation study on the selected DeepLabV3-R34 backbone, focusing on the effect of the training objective. This analysis was designed to verify whether the hybrid loss adopted in the final pipeline is effectively beneficial for this task, beyond the architectural selection itself. 

Since Christmas tree plantation delineation is characterized by severe class imbalance and strong confusion with visually structured non-target classes, the choice of the loss function plays a central role in controlling the precision--recall trade-off. For this reason, after selecting DeepLabV3-R34 as the reference architecture, we compared three alternative loss formulations on this backbone: Tversky, weighted BCE, and the hybrid weighted BCE + Tversky combination. To isolate the effect of the training objective, all experiments were conducted under the same architectural setting, and the final test predictions were binarized using a fixed threshold of 0.70.
The comparison was repeated across multiple seeds to reduce the impact of random initialization, and Table~\ref{tab:loss_ablation} reports the mean and standard deviation of the resulting test metrics. The ablation results show a consistent advantage of the hybrid formulation. Tversky loss alone remains strongly recall-oriented, but yields the weakest overall overlap and precision, indicating that overlap-driven optimization alone is insufficient to control false positives in this highly ambiguous setting. Weighted BCE alone substantially improves IoU, F1-score, and precision, confirming the value of explicit hard-negative weighting. However, the combined weighted BCE + Tversky loss provides the best overall balance, achieving the highest IoU, F1-score, and recall, while also exhibiting the lowest variability across seeds. This indicates that Christmas tree plantation delineation benefits from a hybrid objective that jointly enforces pixel-level discrimination against confusing backgrounds and overlap-aware optimization for the rare target class. For this reason, the hybrid loss was retained for the temporal and large-scale experiments presented in the following.
\subsubsection{Assessment of the proposed HNM strategy}
\label{subsec:hnmnohnm}
\begin{figure}[t]
\centering
\includegraphics[width=\linewidth]{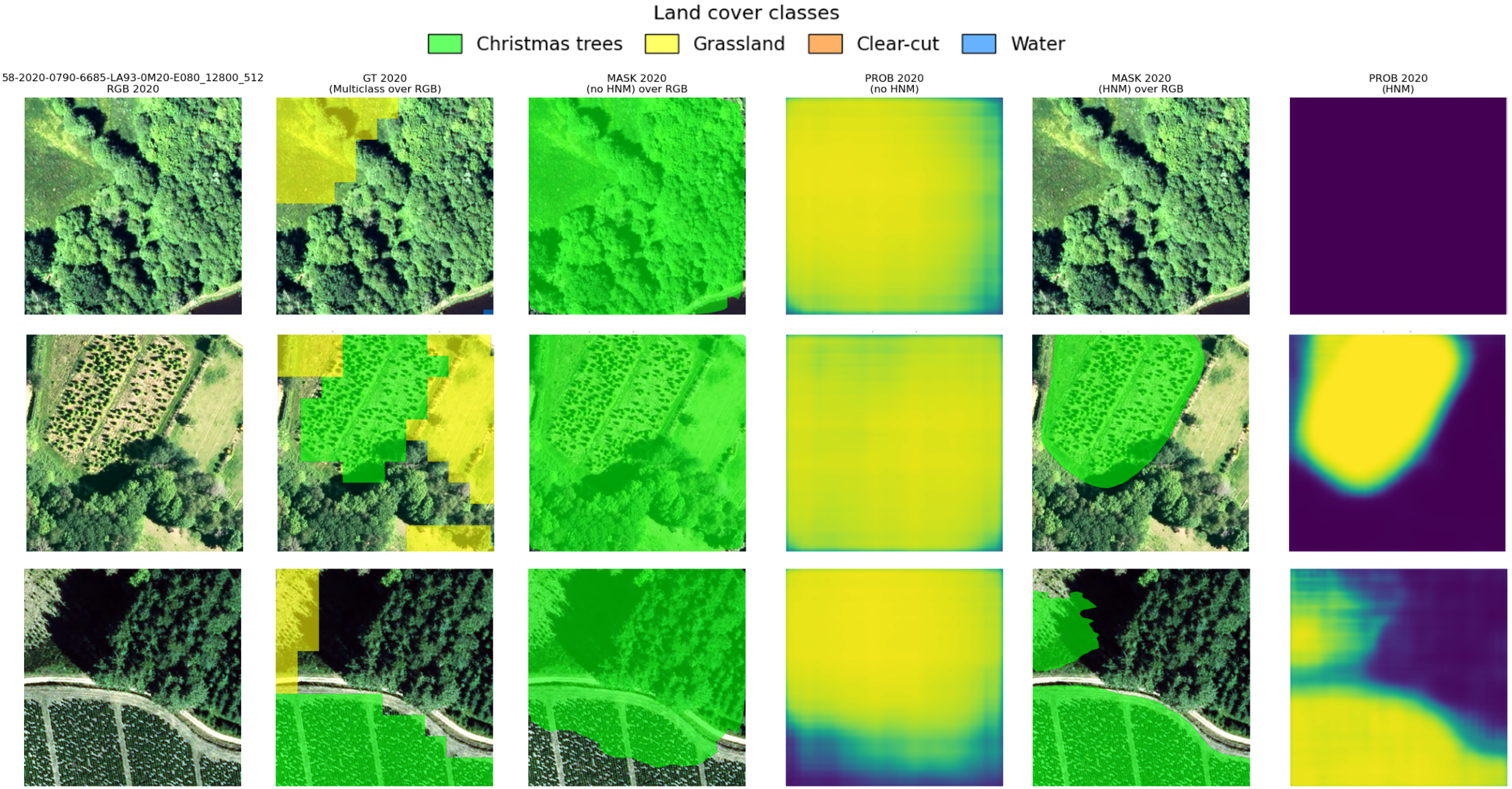}
\caption{Examples of visual comparison for DeepLabV3--R34 trained with and without Hard Negative Mining (HNM) on 2020 test patches. Columns: RGB; GT (multiclass) over RGB; no-HNM mask over RGB; no-HNM $P(\mathrm{CT})$; HNM mask over RGB; HNM $P(\mathrm{CT})$ (shared $[0,1]$ scale). HNM suppresses diffuse false activations and produces more spatially coherent plantation responses.}
\label{fig:hnm_2020_conf}
\end{figure}

We quantify the contribution of HNM pipeline in a controlled setting by comparing the \emph{same} architecture (DeepLabV3--R34) trained \emph{without} HNM against its \emph{HNM-trained} counterpart. More specifically, both experiments share the same DeepLabV3--R34 architecture and the same validation-to-test threshold transfer protocol, while differing in the training setup: the HNM version relies on the proposed hard-negative sampling strategy and hybrid hard-negative-aware loss, whereas the no-HNM baseline is trained with a standard BCE--Dice objective without explicit hard-negative mining.
On the hard-negative 2020 test split, the no-HNM model suffers from a strong \emph{false-positive inflation} (FP $=53{,}474{,}056$), which severely degrades precision ($0.186$) despite a high recall ($0.946$), resulting in poor segmentation quality (F1 $=0.310$, IoU $=0.183$). In contrast, introducing HNM drastically improves background discrimination and false-positive control (FP reduced to $1{,}648{,}335$), yielding a much more balanced precision--recall trade-off (precision $=0.866$, recall $=0.827$) and a large gain in overlap metrics (F1 $=0.846$, IoU $=0.733$). Figure~\ref{fig:hnm_2020_conf} provides qualitative examples consistent with this behavior, highlighting the reduction of spurious activations over visually confusing vegetation patterns. This confirms that the proposed HNM strategy helps the model to discriminate Christmas tree plantations against visually similar but semantically different classes. The proposed HNM strategy contributes in two ways: in the samples selection (exposing the model to patches containing hard negatives rather than random background) and by guiding model's optimization (by weighting the BCE loss according to the similarity between Christmas trees and the selected challenging negative classes).
To complement thresholded metrics, we also report a threshold-free evaluation based on the Precision--Recall (PR) curve. Specifically, we compute the Average Precision (AP), i.e., the area under the PR curve (AUPRC), which is independent of the operating point and therefore not affected by the particular choice of $\tau$. As reported in Table~\ref{tab:hnm_auprc}, AP increases from $0.204$ (no-HNM) to $0.913$ (HNM), confirming that the observed improvement is not merely a consequence of threshold tuning but reflects a substantial increase in separability between Christmas tree plantations and hard negatives. Overall, these results demonstrate that HNM is essential for the highly imbalanced segmentation task of Christmas trees detection.\\

\begin{table}[t]
\centering
\caption{Threshold-free performance on the hard-negative 2020 test split (DeepLabV3--R34). AP corresponds to the area under the Precision--Recall curve (AUPRC).}
\label{tab:hnm_auprc}
\setlength{\tabcolsep}{9pt}
\begin{tabular}{lcc}
\toprule
\textbf{Training regime} & \textbf{AP} & \textbf{AUPRC} \\
\midrule
no-HNM & 0.2042 & 0.2044 \\
\textbf{HNM} & \textbf{0.9125} & \textbf{0.9127} \\
\bottomrule
\end{tabular}
\end{table}
\begin{figure*}[!ht]
\centering
\includegraphics[width=18.2cm]{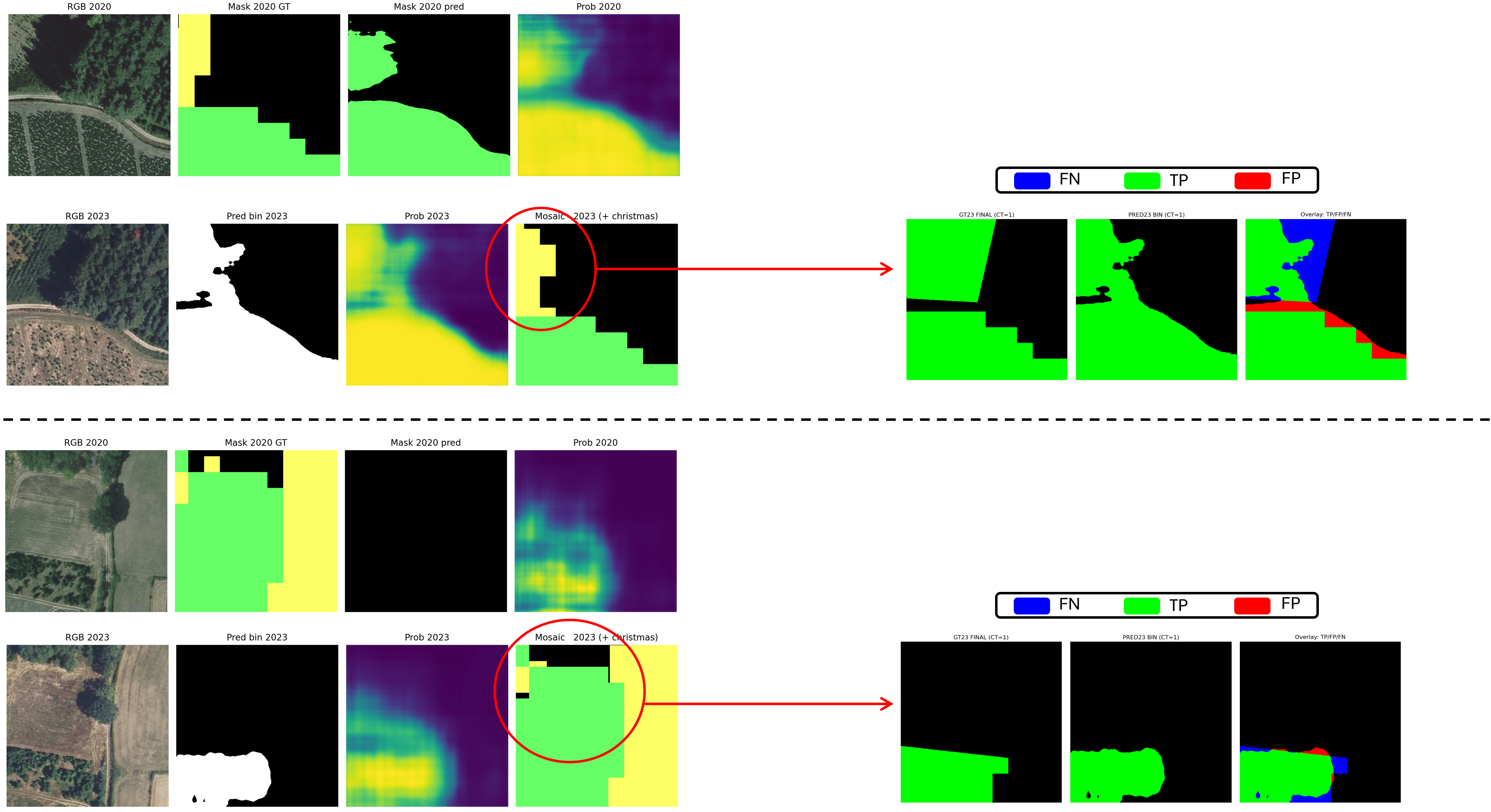}
\captionof{figure}{Targeted refinement examples for 2023 on fixed 2020 test patch locations. For each example (separated by the dashed line), we show the source-year (2020) RGB and labels, the target-year (2023) RGB and model outputs, the proxy mosaic used as initial labels, and the final corrected GT. The refinement resolves systematic proxy errors (missing plantations and boundary/semantic mismatches) while preserving unchanged areas.}
\label{fig:qual_corrections_2023}
\vspace{1.2em}
\includegraphics[width=19cm]{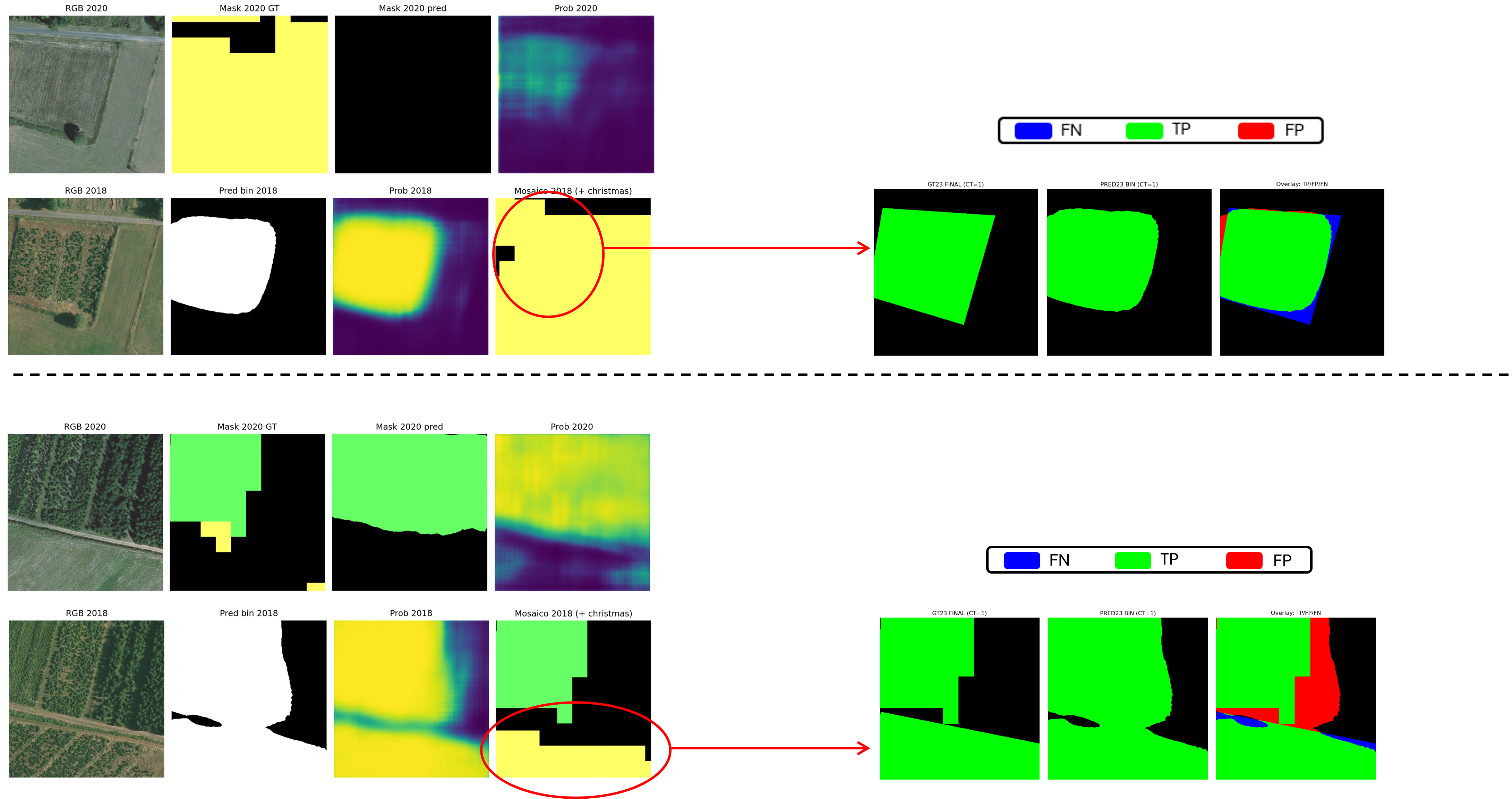}
\captionof{figure}{Examples of corrected 2017/2018 annotations using the same targeted refinement workflow adopted for 2023. The procedure focuses on patches showing clear inconsistencies between target-year RGB evidence, model predictions, and the proxy mosaic, producing final labels with improved spatial consistency for evaluation.}
\label{fig:qual_corrections_1718}
\end{figure*}

\subsection{Temporal Inference on 2017/2018 and 2023}
\label{subsec:temporal test}
We extend the analysis of our model to other years, to evaluate the temporal generalization of the model. We first conduct a study on the same spatial scale as the 2020 test set, where we extended the GT enabling a precise assessment at the pixel level, before deploying the model at the scale of the entire Morvan Regional National Park.

\subsubsection{Small-scale analysis}
We deploy our DeepLabV3--R34 trained with HNM and the selected hybrid training objective to RGB imagery acquired in 2017/2018 and 2023 over the same spatial patch locations used in the 2020 test split. This design isolates temporal effects (radiometry, phenology, acquisition conditions, and landscape evolution) while keeping the evaluation footprint fixed.
A central challenge is that, unlike 2020, no GT for Christmas tree plantations was initially available for 2017/2018 and 2023. The selection of this spatial scale enabled to establish a protocol to build a pixel-level GT for the considered years. The proposed protocol works as follows. To avoid re-annotating the full dataset, we first built a \emph{proxy} GT for each target year by \textit{(i)} transferring the 2020 Christmas tree parcels under a temporal stability assumption and \textit{(ii)} integrating auxiliary land-cover layers (grassland, clear-cut, water) available for the target year to filter out parcels that did not contain Christmas trees. While operationally convenient, this proxy can introduce systematic errors (missing plantations, boundary inaccuracies, and semantic mismatches), which leads to systematic label noise.
To reduce proxy-induced bias while keeping manual effort bounded, we adopted a targeted, visually guided refinement workflow. Each patch was inspected through overlays of: \textit{(i)} target-year RGB imagery, \textit{(ii)} model probability maps and binary predictions, and \textit{(iii)} the proxy GT. Only patches showing clear inconsistencies were selected for correction.
Edits were performed interactively (Napari) using polygon-based annotations to ensure clean and reproducible boundaries. Two explicit operations were defined:
\textbf{ADD}, to insert missing Christmas tree parcels (proxy false negatives), and \textbf{REMOVE}, to delete falsely labeled parcels (proxy false positives). The refined GT is constructed as:
\begin{equation}
\mathrm{GT}_{\mathrm{final}} = \left(\mathrm{GT}_{\mathrm{proxy}} \cup \mathrm{ADD}\right) \setminus \mathrm{REMOVE}.
\end{equation}
Importantly, the original proxy labels were \emph{never overwritten}; corrected masks were stored separately and evaluation metrics were recomputed using the corrected GT when available, otherwise falling back to the proxy labels. This enables a direct \emph{before/after} comparison while preserving full traceability of the annotation process. Qualitative examples of corrected patches are reported for 2023 and for 2017/2018 (Fig.~\ref{fig:qual_corrections_2023}--\ref{fig:qual_corrections_1718}).
From a quantitative perspective, our re-annotation strategy led to correcting a limited subset of visually inconsistent patches (22 out of 258, i.e. $\sim$8.5\%,  for 2023 and only 8 for 2017/18), reducing proxy label errors with minmal annotation effort. The qualitative overlays confirm that corrections mostly address missing plantation regions and boundary mismatches in the proxy GT (Fig.~\ref{fig:qual_corrections_2023}). Although the temporal evaluation is defined on the same spatial patch locations as the 2020 test split, the effective evaluation set is not identical across years. In particular, for 2017/2018 some RGB tiles were unavailable or invalid in the source geoportail because the acquisition campaign did not fully cover all locations. As a result, a subset of the 2020 test patches could not be evaluated for 2017/2018, and the corresponding temporal test area is smaller than for 2023. Qualitative examples illustrate how the refinement resolves missing or misaligned plantation boundaries and clarifies ambiguous areas in the proxy mosaic (Fig.~\ref{fig:qual_corrections_1718}).

The quantitative assessment on the refined GT is reported in Table \ref{tab:temporal_proxy_vs_refined}. All reported metrics fall in a similar range as for 2020, demonstrating that the model generalizes to other years and can be deployed to map of Christmas tree plantations over time.

\begin{table}[t]
\centering
\caption{Temporal inference results on the refined GT on the fixed 2020 test patch locations. Metrics are computed for the HNM pipeline.}
\label{tab:temporal_proxy_vs_refined}
\setlength{\tabcolsep}{6pt}
\begin{tabular}{l c c c c}
\toprule
\textbf{Year} & \textbf{IoU $\uparrow$} & \textbf{F1 $\uparrow$} & \textbf{Precision $\uparrow$} & \textbf{Recall $\uparrow$} \\
\midrule
2023 & \textbf{0.691} & {0.817} & {0.876} & \textbf{0.766} \\
\midrule
2017/2018  & {0.751} & {0.858} & {0.893} & {0.826} \\
\bottomrule
\end{tabular}
\end{table}

\subsubsection{Large-scale detection of Christmas tree plantations}
\label{subsec:large test}

To assess the operational scalability of the proposed framework, we performed large-scale detection over the common spatial extent covered by the 2017/2018 imagery. Predicted binary masks were converted into polygon features, and both the 2023 predictions and the 2023 reference layer were spatially cropped to the area jointly covered by the multi-temporal mosaics. This ensured that all subsequent analyses were carried out over a common spatial domain.

This validation setting also highlights the intrinsic difficulty of the problem. Over the common evaluation extent, corresponding to 87,309.4 ha, the 2023 GT plantations occupy only 1,782.2 ha, i.e., 2.04\% of the total area, whereas the remaining 97.96\% corresponds to background. The resulting background-to-positive area ratio is therefore approximately 48:1, confirming that the large-scale task is severely imbalanced. In this setting, the objective is not to map a dominant land-cover class, but rather to detect sparse and spatially limited plantation patches over a very large predominantly negative landscape.

Under fixed validation conditions at 5 m spatial resolution and without additional AOI shrinkage, a targeted post-processing step was applied to refine the raw predictions. The selected configuration consisted of removing connected components smaller than 900 m$^2$ using 4-connectivity, without hole filling and without erosion. This choice was retained as the least invasive refinement of the raw predictions, providing conservative denoising while preserving the overall detection capacity of the baseline output.
\begin{figure*}[!ht]
    \centering
    \subfloat[Evaluation year 2017/2018]{
        \includegraphics[width=15cm]{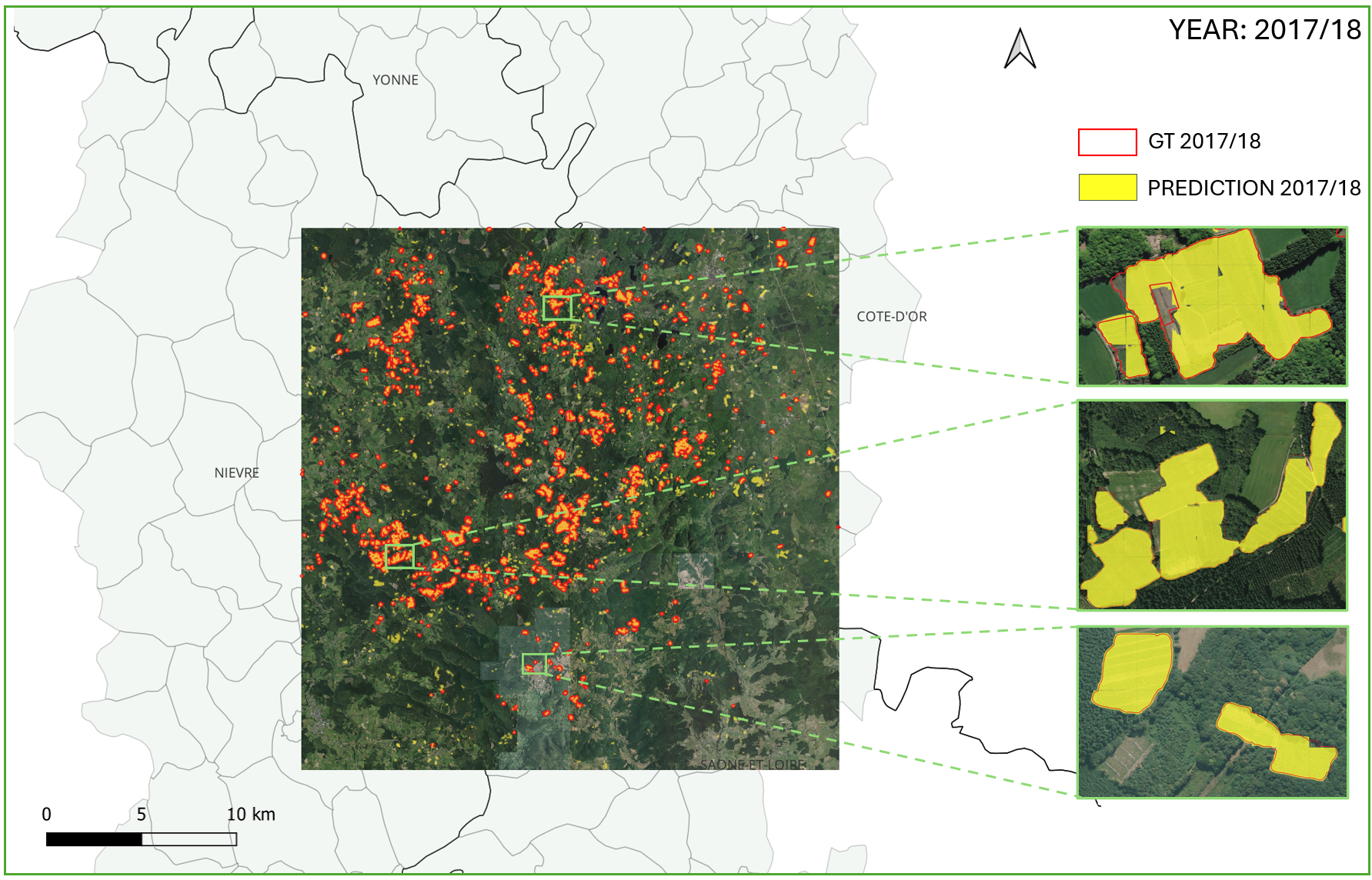}
        \label{fig:large_scale_2017}
    }
    
    \vspace{0.4cm} 
    
    \subfloat[Evaluation year 2023]{
        \includegraphics[width=15cm]{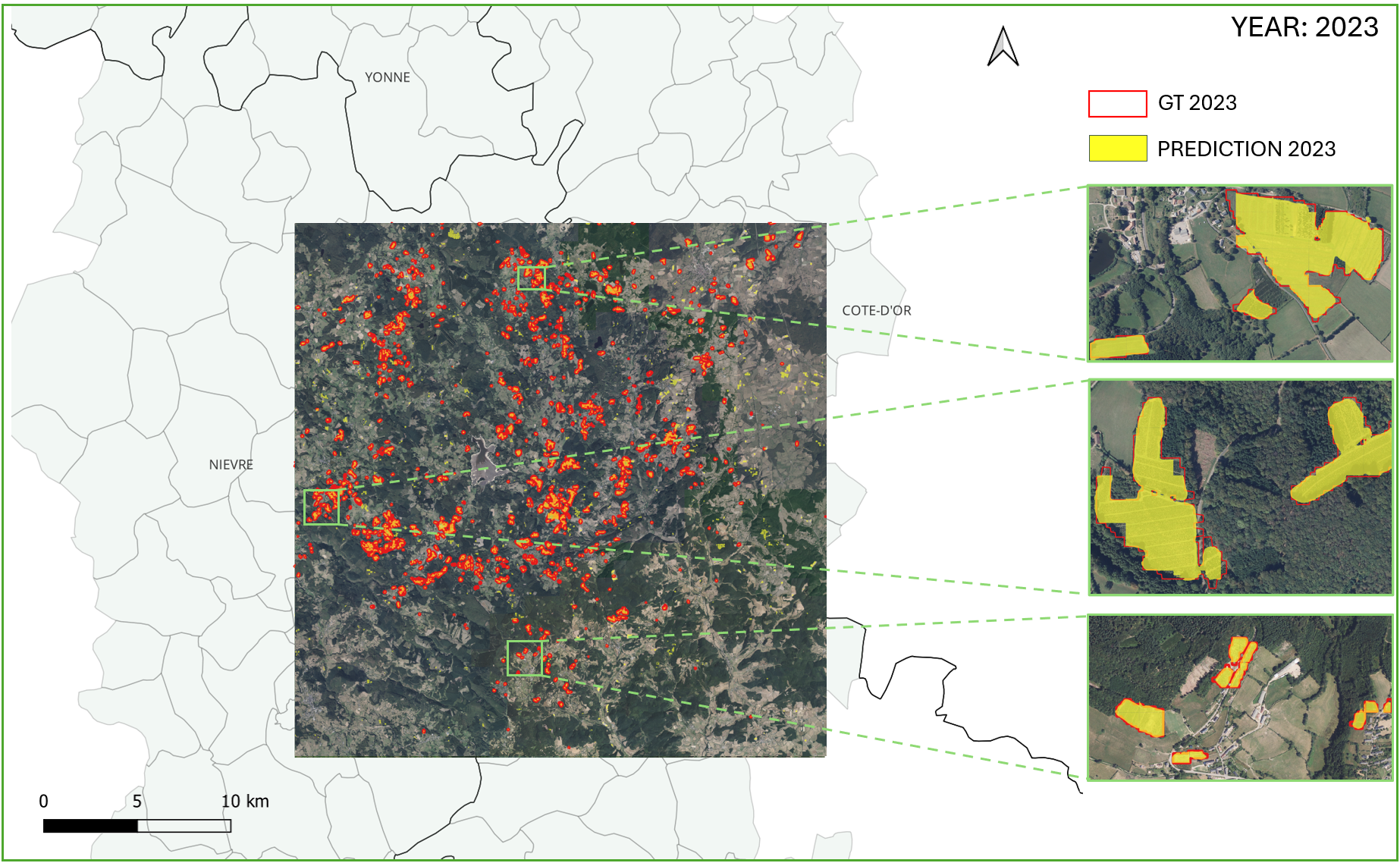}
        \label{fig:large_scale_2023}
    }
    
    \caption{Qualitative evaluation of the large-scale detection framework over the Morvan Regional Natural Park for the 2017/2018 and 2023 evaluation years. The main panels display the wall-to-wall spatial distribution of the predicted Christmas tree plantations against their respective ground-truth inventories across the entire 87,000 ha extent. The side panels provide representative zoomed-in areas detailing the local matching behavior: solid yellow masks indicate the model predictions, while red boundaries denote the reference ground-truth polygons. The selected insets highlight the model's accuracy and temporal consistency across different spatial contexts, including dense central clusters and isolated peripheral regions.}
    \label{fig:large_scale_combined}
\end{figure*}

For the 2023 common extent, the final post-processed prediction achieved precision = 0.653, recall = 0.736, F1-score = 0.692, and IoU = 0.529, with a false-positive area of 695.81 ha and a false-negative area of 471.39 ha. For the 2017/2018 common extent, the same post-processing configuration yielded precision = 0.597, recall = 0.879, F1-score = 0.711, and IoU = 0.551, with a false-positive area of 889.80 ha and a false-negative area of 181.72 ha. Table~\ref{tab:large_scale_final} summarizes the reference plantation area, predicted area, and final area-based validation results over the common study extent. Notably, the selected pipeline recovers 73.6\% of the reference plantation area in 2023 and 87.9\% in 2017/2018 over the common validation extent.

\begin{table}[t]
\centering
\caption{Final large-scale area-based validation over the common study extent after the selected post-processing step.}
\label{tab:large_scale_final}
\begin{tabular*}{\columnwidth}{@{\extracolsep{\fill}}lcc@{}}
\hline
Metric & 2023 & 2017/2018 \\
\hline
GT area (ha)    & 1,782.21 & 1,498.44 \\
Pred. area (ha) & 2,006.63 & 2,206.52 \\
FP (ha)         & 695.81  & 889.80 \\
FN (ha)         & 471.39  & 181.72 \\
Precision       & 0.653  & 0.597 \\
Recall          & 0.736  & 0.879 \\
F1-score        & 0.692  & 0.711 \\
IoU             & 0.529  & 0.551 \\
\hline
\end{tabular*}
\end{table}

Overall, the large-scale validation confirms that the proposed pipeline can recover a substantial portion of plantation area over a very large and highly imbalanced study domain. At the same time, the remaining errors are consistent with the challenges highlighted in the previous experiments, namely the presence of visually confusing background patterns that can still generate residual false-positive fragments.
To complement these quantitative metrics, Fig.~\ref{fig:large_scale_combined} provides a visual representation of the large-scale deployment for both the 2017/2018 and 2023 evaluation years. A qualitative inspection confirms the model’s ability to locate Christmas tree parcels far beyond the primary production clusters, successfully identifying isolated plantations at the periphery of the Natural Park across both time steps. As shown in the detailed inset maps, the predicted masks closely adhere to the geometric boundaries of the active parcels. The visual overlays confirm that while some false positives are occasionally generated, primarily over recently disturbed forest patches or similarly structured herbaceous clearings, the wall-to-wall maps demonstrate the strong operational viability of the proposed framework. A deeper geospatial and agronomic analysis regarding the long-term landscape evolution of these parcels is left for future studies.

\section{Conclusion and Discussion}
\label{sec:conclusion}

This study shows that Christmas tree plantation delineation from high-resolution RGB orthophotography should be treated as a distinct RS task rather than as a straightforward extension of generic forest or orchard mapping. The difficulty of the problem lies not only in the small spatial footprint of the target class, but also in its strong visual similarity to other managed or transitional land covers, especially grasslands, clear-cuts, and heterogeneous vegetated backgrounds. In this context, the proposed framework proved effective because it was designed explicitly for rare target detection under severe class imbalance and hard negative confusion. In particular, the comparison with the standard training configuration confirms that HNM pipeline is not a secondary refinement, but a central component of the pipeline, substantially improving the rejection of visually plausible non-target patterns and making the model more suitable for operational deployment. The experiments further indicate that the effectiveness of the framework does not depend solely on architectural complexity, but on the balance between contextual representation, discrimination capacity, and post-processing design. DeepLabV3 with a ResNet-34 backbone emerged as the most reliable configuration, providing the strongest compromise on the 2020 benchmark and maintaining robust behavior under transfer to 2023 and 2017/2018 imagery. This temporal consistency is particularly relevant because the model was trained on a single annotated reference year, yet remained capable of detecting plantation areas across different acquisition periods and landscape conditions. The large-scale validation strengthens this result by showing that the pipeline preserves useful detection capacity even over a very large and highly imbalanced study domain. At the same time, it also clarifies the main residual limitation of the method, namely the persistence of false-positive fragments in visually structured background areas that partially reproduce the spatial organization of plantation parcels. \\
All these results show that the proposed approach provides a credible first operational baseline for large-scale Christmas tree plantation mapping from RGB aerial imagery. 
From an application perspective, such an objective mapping tool is highly anticipated to inform local land-use debates. In the Morvan, environmental organisations and local collectives have pointed to the effects of monoculture plantations and agrochemical use on soil quality, water resources, and biodiversity. At the same time, producers highlight the economic importance of the sector and their ongoing efforts to adopt more sustainable practices, for instance through eco-responsibility labels and protected geographical indication schemes. By providing reliable, spatially explicit information on where these plantations are located and how they evolve over time, our framework offers a neutral basis for environmental assessment, land-use planning, and socio-political analysis.
Building on this operational baseline, the study opens several directions for further investigation. A first perspective concerns the joint analysis of multi-temporal layers to better characterize how plantation areas evolve, aiming to describe not only where Christmas tree parcels are located but also how land is progressively transformed and reused across production cycles. In this respect, the construction of refined GT information for multiple dates provides an important starting point for moving beyond single-date detection toward a more explicit interpretation of plantation dynamics and land-use trajectories. Another important extension concerns the explicit use of temporal information within the learning process itself, for instance through multi-date training or change-aware modeling, which could exploit plantation growth and harvesting dynamics more directly than the current single-reference-year setting. Finally, object-aware or instance-level formulations remain an open direction of exploration, as they could help reduce residual fragmentation and improve parcel-level delineation in visually ambiguous areas, while broader transfer experiments across different production regions would be essential to assess the geographical generalizability of the proposed framework.

\addtolength{\textheight}{-0.9cm} 

\section*{Aknowledgements}
This work was partially supported by the French National Research Agency in the framework of the two "France 2030" programs: ANR-15-IDEX-0002 through the LabEx ITTEM and the ANR-23-IACL-0006.

\bibliographystyle{IEEEtran}
\bibliography{bib.bib}
 


\if 


    \begin{IEEEbiography}[{\includegraphics[width=1in,height=1.15in,clip,keepaspectratio]{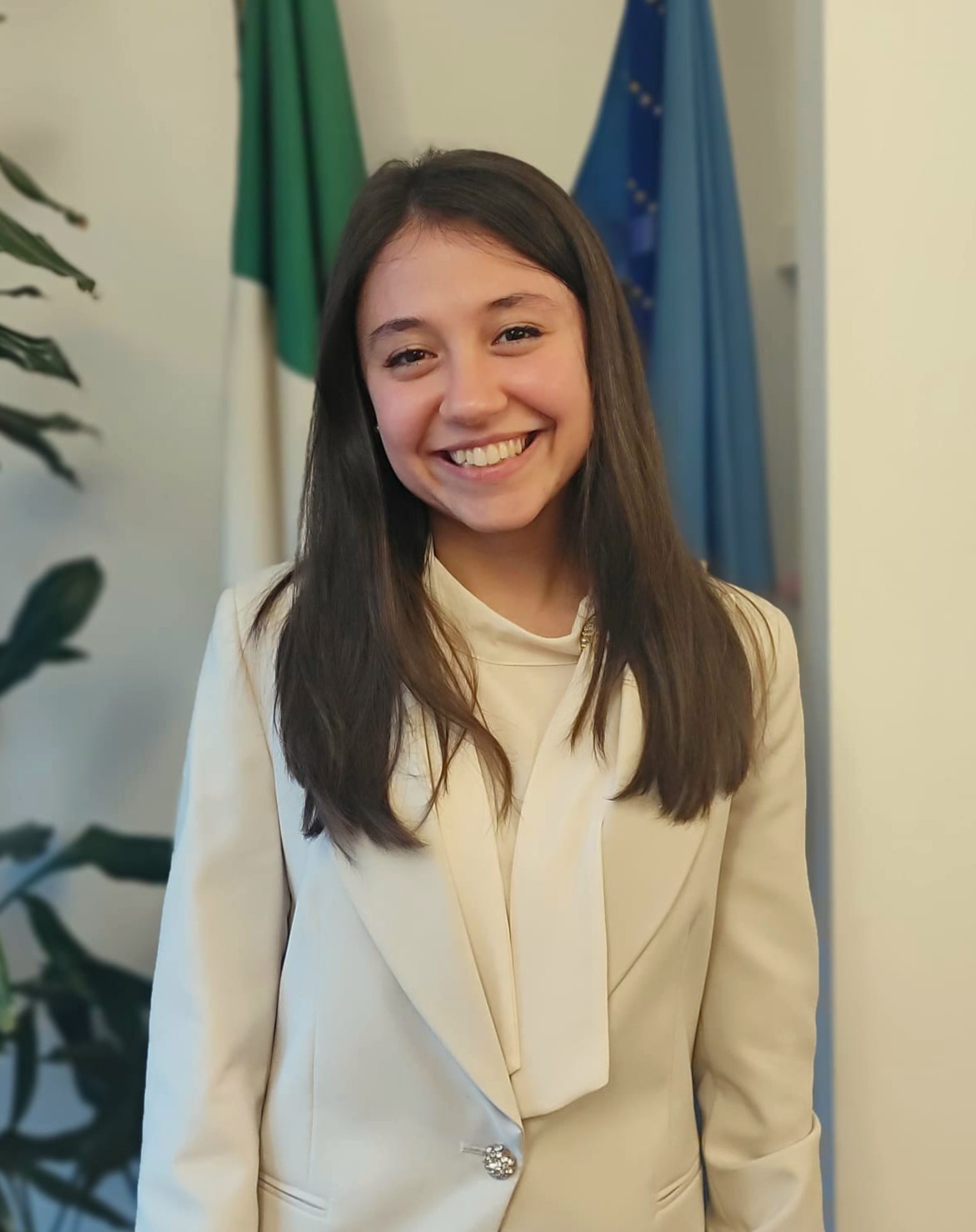}}]{Francesca Razzano}
    graduated cum laude in Electronic Engineering for Automation and Telecommunications from the University of Sannio in 2023. She is currently enrolled in the Ph.D. program in Information and Communication Technology and Engineering at the University of Parthenope, in Naples, under the supervision of Prof. Gilda Schirinzi and co-supervision of Prof. Silvia L. Ullo. Her research focuses primarily on Remote Sensing and satellite data analysis, as well as the application of Artificial Intelligence techniques for Earth observation. In particular, she investigates water quality monitoring and forest tree height estimation. She has also contributed to research on the fusion of optical and SAR data for different tasks. In addition, she works on the development of onboard AI for Remote Sensing systems. Her professional experience includes a position as a Visiting Researcher at the European Space Agency’s $\Phi$-Lab. She has co-authored papers and articles presented at renowned conferences in the field of remote sensing. She is an IEEE Student Member, actively involved in IEEE GRSS IDEA initiatives, and participates in the IEEE Young Professionals Affinity Group of the Italy Section.
    \end{IEEEbiography}

    \begin{IEEEbiography}[{\includegraphics[width=1in,height=1.20in,clip,keepaspectratio]{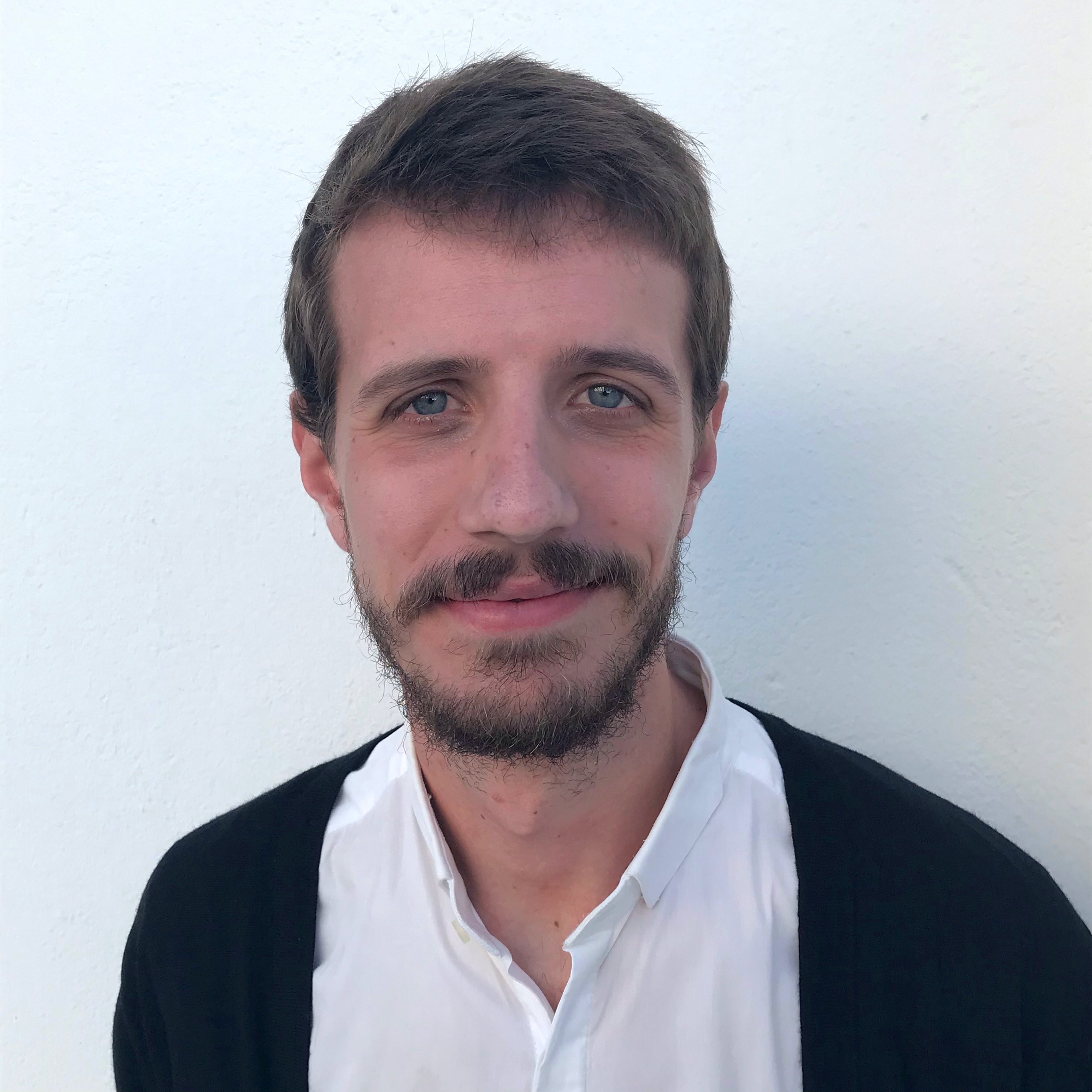}}]{Emanuele Dalsasso}
    Emanuele Dalsasso received the M.Sc. degree in information and communication engineering (\textit{summa cum laude}) from the University of Trento, Italy, in 2018 and the Ph.D. degree in 2022 from Télécom Paris, Institut Polytechnique de Paris in 2022 with a thesis on "Deep Learning for SAR Imagery: from denoising to scene understanding". 
    He is currently a Research Scientist with the Centre Inria de l'Université Grenoble Alpes, Grenoble, France. His research interests include remote sensing image analysis, in particular Synthetic Aperture Radars, machine learning and deep learning. He was a recipient of the IEEE GRSS Symposium Prize Paper Award in 2021 for his works on speckle reduction using deep learning.
    \end{IEEEbiography}

    \begin{IEEEbiography}[{\includegraphics[width=1.2in,height=1.15in,clip,keepaspectratio]{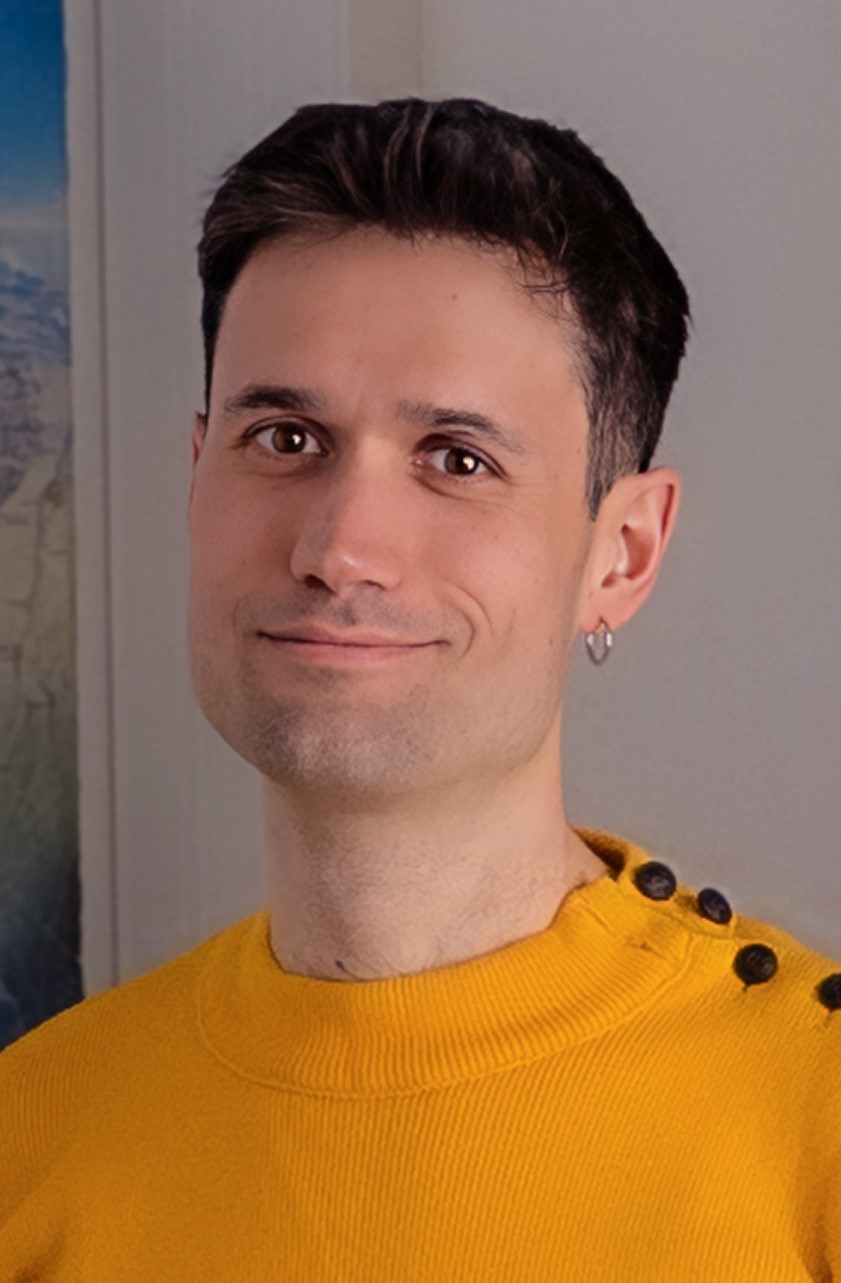}}]{Adrien Baysse-Lainé}
    received the M.Sc. degree in environmental geography from the Paris Cité University (Paris, France) in 2014, and in general geography from the Ecole normale supérieure Ulm (Paris, France) in 2015; then the Ph.D. degree in rural geography from the University of Lyon (Lyon, France) in 2018. Since 2020, he is a Researcher with the French National Centre for Scientific Research (CNRS), within the Pacte Social Sciences Research
Centre (Grenoble, France). His research interests include power relationships and dynamics of inequities around farmland control and access, inclusion of soil quality criteria in spatial planning and regional governance, controversies on tree plantations. In 2023-2024, he was Co-Director of the International Geography festival of Saint-Dié (France). Since 2025 he is the Co-Editor-in-Chief of the Journal of Alpine Resarch. In 2026, his achievements were rewarded with the CNRS Bronze Medal (early-career French scientific honour).
    \end{IEEEbiography}

    \begin{IEEEbiography}[{\includegraphics[width=1in,height=1.15in,clip,keepaspectratio]{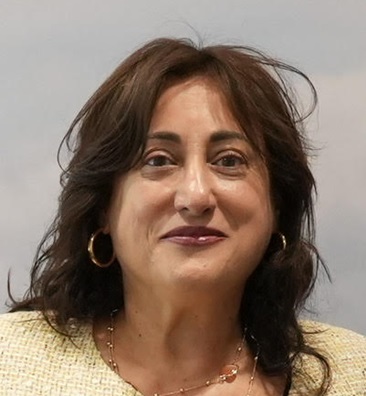}}]{Silvia Liberata Ullo}
IEEE Senior Member, IEEE AESS Italy Chapter Chair, IEEE GRSS Europe Liaison and AdCom Member, Member of the Image Analysis and Data Fusion Technical Committee of the IEEE GRSS since 2020 and Chair of the IADF MIA Working Group, Industry Liaison for IEEE Joint ComSoc/VTS Italy Chapter. National Referent for FIDAPA BPW Italy Science and Technology Task Force (2019-2021). 
Graduated with Laude in Electronic Engineering with a specialization in Telecommunications at Federico II University in Naples (Italy), in 1989. She pursued a Master of Science in Management at the Massachusetts Institute of Technology (MIT) in Cambridge (U.S.A.) in 1992. Researcher since 2004 at the University of Sannio, Benevento (Italy), and Associate Professor since 2026. Member of the Academic Senate (until October 2025) and PhD Professors’ Board (present). Courses: Probability and Signals, Geospatial Data Analysis (Bachelor program); Earth monitoring and mission analysis Lab (Master program), Optical and radar Remote Sensing (Ph.D. program).  Authored 140+ research papers, co-authored many book chapters and served as editor of two books. Associate Editor of relevant journals (IEEE TGRS, JSTARS, GRSL, MDPI Remote Sensing, Springer Arabian Journal of Geosciences and others). Co-Editor-in-Chief of IET Image Processing. Guest Editor of many special issues. Research interests: signal processing, radar systems, sensor networks, smart grids, remote sensing, satellite data analysis, machine learning and quantum ML applied to remote sensing.
    \end{IEEEbiography}
\vspace{-4cm}
    \begin{IEEEbiography}[{\includegraphics[width=1in,height=1.15in,clip,keepaspectratio]{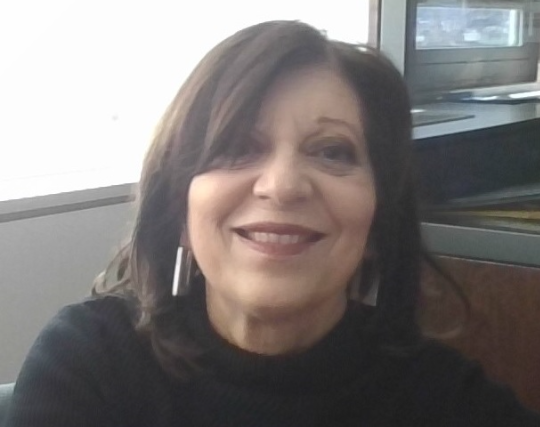}}]{Gilda Schirinzi}
    graduated cum laude in electronic engineering at the University of Naples "Federico II." From 1985 to 1986, she was at the European Space Agency, ESTEC, Noordwijk, The Netherlands. In 1988, she joined the Istituto di Ricerca per l'Elettromagnetismo e i Componenti Elettronici (IRECECNR), Naples, Italy. In 1998, she joined the University of Cassino, Italy, as an associate professor of telecommunications, and in 2005, she became a full professor. Since 2008, she has been at the University of Naples “Parthenope.” Her main scientific interests are in the field of signal processing for Remote Sensing applications, with particular reference to synthetic aperture radar (SAR) interferometry and tomography. She is a Senior Member of the IEEE.
    \end{IEEEbiography}
\vspace{5cm}
    \begin{IEEEbiography}[{\includegraphics[width=1in,height=1.15in,clip,keepaspectratio]{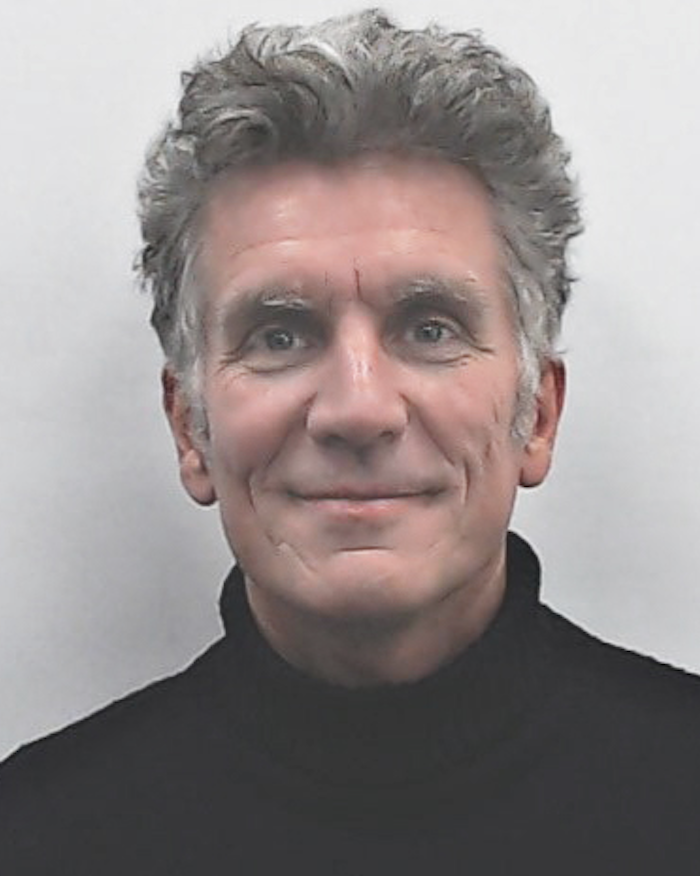}}]{Jocelyn Channusot}
    (M’04–SM’04–F’12) received the M.Sc. degree in electrical engineering
    from the Grenoble Institute of Technology (Grenoble INP), Grenoble, France, in 1995, and
    the Ph.D. degree from the Université de Savoie, Annecy, France, in 1998. From 1999 to 2023,
    he has been with Grenoble INP, where he was a Professor of signal and image processing. He
    is currently a Research Director with INRIA, Grenoble. His research interests include image
    analysis, hyperspectral remote sensing, data fusion, machine learning and artificial
    intelligence. He has been a visiting scholar at Stanford University (USA), KTH (Sweden) and
    NUS (Singapore). Since 2013, he is an Adjunct Professor of the University of Iceland. In
    2015-2017, he was a visiting professor at the University of California, Los Angeles (UCLA).
    He holds the AXA chair in remote sensing and is an Adjunct professor at the Chinese
    Academy of Sciences, Aerospace Information research Institute, Beijing.
    Dr. Chanussot is the founding President of IEEE Geoscience and Remote Sensing French
    chapter (2007-2010) which received the 2010 IEEE GRS-S Chapter Excellence Award. He
    has received multiple outstanding paper awards. He was the Vice-President of the IEEE
    Geoscience and Remote Sensing Society, in charge of meetings and symposia (2017-2019).
    He was the General Chair of the first IEEE GRSS Workshop on Hyperspectral Image and
    Signal Processing, Evolution in Remote sensing (WHISPERS). He was the Chair (2009-2011)
    and Cochair of the GRS Data Fusion Technical Committee (2005-2008). He was a member
    of the Machine Learning for Signal Processing Technical Committee of the IEEE Signal
    Processing Society (2006-2008) and the Program Chair of the IEEE International Workshop
    on Machine Learning for Signal Processing (2009). He is an Associate Editor for the IEEE
    Transactions on Geoscience and Remote Sensing, the IEEE Transactions on Image Processing
    and the Proceedings of the IEEE. He was the Editor-in-Chief of the IEEE Journal of Selected
    Topics in Applied Earth Observations and Remote Sensing (2011-2015). In 2014 he served as
    a Guest Editor for the IEEE Signal Processing Magazine. He is a Fellow of the IEEE, an
    ELLIS Fellow, a Fellow of the Asia-Pacific Artificial Intelligence Association, a member of
    the Institut Universitaire de France (2012-2017) and a Highly Cited Researcher (Clarivate
    Analytics/Thomson Reuters, since 2018).
    \end{IEEEbiography}
\end{document}